\documentclass[10pt,twocolumn,letterpaper]{article}

\usepackage{cvpr}              %

\usepackage[dvipsnames, table]{xcolor}
\usepackage{makecell}

\definecolor{gold}{rgb}{1.0, 0.87, 0.0}
\definecolor{silver}{rgb}{0.75, 0.75, 0.75}
\definecolor{bronze}{rgb}{0.8, 0.5, 0.2}

\newcommand\method{GoMAvatar\xspace}

\usepackage{multirow}
\usepackage{bm}
\usepackage{amssymb}
\usepackage{adjustbox}
\usepackage{makecell}
\usepackage{subcaption}
\usepackage{graphicx}
\usepackage{gensymb}

\makeatletter
\usepackage{xspace}
\def\@onedot{\ifx\@let@token.\else.\null\fi\xspace}
\DeclareRobustCommand\onedot{\futurelet\@let@token\@onedot}

\def\eg{\emph{e.g}\onedot} 
\def\ie{\emph{i.e}\onedot}

\newcommand{\figref}[1]{Fig\onedot~\ref{#1}}
\newcommand{\equref}[1]{Eq\onedot~\eqref{#1}}
\newcommand{\secref}[1]{Sec\onedot~\ref{#1}}
\newcommand{\tabref}[1]{Tab\onedot~\ref{#1}}

\newcommand{\ignorethis}[1]{}

\makeatletter
\DeclareRobustCommand\onedot{\futurelet\@let@token\@onedot}
\def\@onedot{\ifx\@let@token.\else.\null\fi\xspace}

\def\eg{\emph{e.g}\onedot} 
\def\ie{\emph{i.e}\onedot}

\makeatother

\definecolor{citecolor}{RGB}{34,139,34}
\definecolor{mydarkblue}{rgb}{0,0.08,1}
\definecolor{mydarkgreen}{rgb}{0.02,0.6,0.02}
\definecolor{mydarkred}{rgb}{0.8,0.02,0.02}
\definecolor{mydarkorange}{rgb}{0.40,0.2,0.02}
\definecolor{mypurple}{RGB}{111,0,255}
\definecolor{myred}{rgb}{1.0,0.0,0.0}
\definecolor{mygold}{rgb}{0.75,0.6,0.12}
\definecolor{myblue}{rgb}{0,0.2,0.8}
\definecolor{mydarkgray}{rgb}{0.66,0.66,0.66}

\usepackage{booktabs}

\definecolor{cvprblue}{rgb}{0.21,0.49,0.74}
\usepackage[pagebackref,breaklinks,colorlinks,citecolor=cvprblue]{hyperref}

\usepackage[accsupp]{axessibility} %

\def\paperID{10222} %
\def\confName{CVPR}
\def\confYear{2024}

\title{\method: Efficient Animatable Human Modeling from Monocular Video Using Gaussians-on-Mesh}

\author{
Jing Wen 
\quad Xiaoming Zhao
\quad Zhongzheng Ren
\quad Alexander G. Schwing
\quad Shenlong Wang \\
University of Illinois Urbana-Champaign \\
{\tt\small \{jw116, xz23, zr5, aschwing, shenlong\}@illinois.edu}\\
\small {\url{https://wenj.github.io/GoMAvatar/}}
}

\begin{document}

\twocolumn[{
\renewcommand\twocolumn[1][]{#1}
\maketitle
\vspace{-10mm}
\includegraphics[width=0.99\textwidth,trim=0 2cm 7cm 8cm,clip]{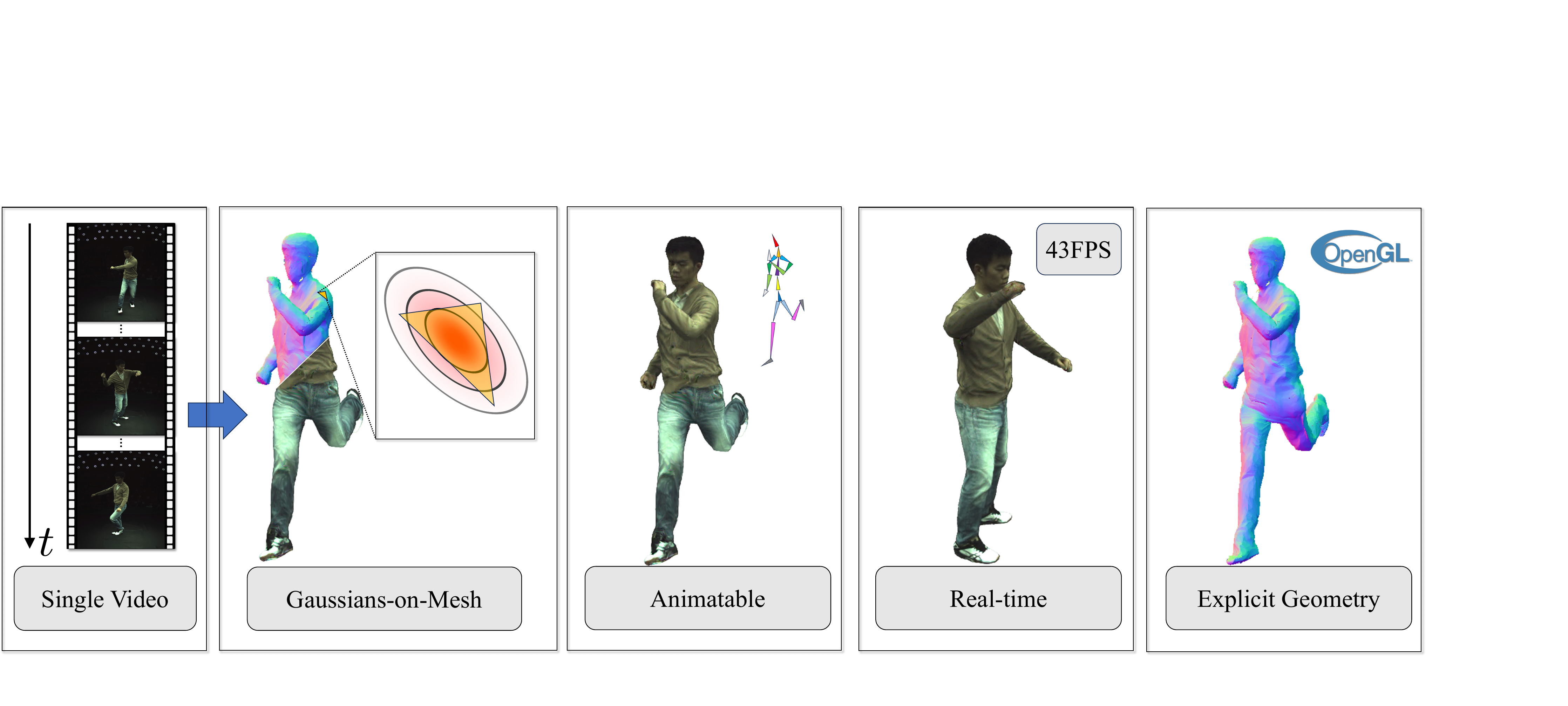}
\vspace{-3mm}
\captionof{figure}{\method takes a monocular RGB video (left) as input to establish an explicit and accurate 4D representation of a dynamic human. It can render efficiently at novel views and poses with state-of-the-art quality. Additionally, it is extremely compact ({\bf 3.63} MB per subject), efficient ({\bf 43} FPS), and seamlessly compatible with the graphics pipeline such as OpenGL.}
\label{fig:teaser}
\vspace{3mm}
}]

\begin{abstract}
We introduce \method, a novel approach for real-time, memory-efficient, high-quality animatable human modeling. \method takes as input a single monocular video to create a digital avatar capable of re-articulation in new poses and real-time rendering from novel viewpoints, while seamlessly integrating with rasterization-based graphics pipelines. Central to our method is the \textbf{Gaussians-on-Mesh (GoM)} representation, a hybrid 3D model combining rendering quality and speed of Gaussian splatting with geometry modeling and compatibility of deformable meshes. We assess \method on ZJU-MoCap, PeopleSnapshot, and various YouTube videos. \method~matches or surpasses current monocular human modeling algorithms in rendering quality and significantly outperforms them in computational efficiency (43 FPS) while being memory-efficient (3.63 MB per subject).
\end{abstract}

\section{Introduction}
\label{sec:intro}

High-fidelity, animatable digital avatar modeling is crucial for various applications such as movie making, healthcare, AR/VR, and simulation. Conventional approaches carried out in Motion Capture (MoCap) studios are slow, expensive, and cumbersome, due to costly wearable devices~\cite{loper2014mosh, park2006capturing} and intricate multi-view camera systems~\cite{Joo_2017_TPAMI,wuu2022multiface}. Hence, to enable widespread personal use, affordable methods which only rely on monocular RGB videos for creating digital avatars are much desired.

Reconstruction of digital humans from monocular videos has been studied intensively recently~\cite{Geng2023LearningNV, Jiang2022SelfReconSR, Su2021ANeRFAN, Weng2022HumanNeRFFR, yu2023monohuman}. The key lies in choosing a suitable 3D representation, flexible for articulation, efficient for rendering and storage, and capable of modeling high-quality geometry and appearance all while being easily integrated into graphics pipelines.
Despite various proposals, no animated 3D representation has met all these needs. 
Neural fields based avatars~\cite{Weng2022HumanNeRFFR, yu2023monohuman, Geng2023LearningNV, jiang2022neuman} offer photorealism, but they are challenging to articulate and lack explicit geometry, making them less compatible with game engines. 
Mesh-based methods~\cite{Rocco2023RealtimeVR} excel in articulation and rendering but fall short in modeling topological changes and high-quality appearance.
Point-based methods~\cite{Zheng2022PointAvatarDP} are limited by incomplete topology and surface geometry. 
Recent successes of Gaussian splatting in neural rendering motivate extensions to free-form dynamic scenes~\cite{wu20234dgaussians}, but a knowledge gap exists in {\it how to leverage Gaussians for articulatable humans}. 
Besides, the lack of explicit surface modeling of Gaussian splats hinders their broader use in digital avatar modeling. 

To address these challenges, we present \method, a novel digital avatar modeling framework. \method~operates on a single monocular video and yields an articulated character that encodes high-fidelity appearance and geometry. It is both articulable and memory-efficient, rendering in real-time (see \figref{fig:teaser}). Central to the framework is a novel articulated human representation, which we refer to as \emph{Gaussians-on-Mesh} (GoM) (\secref{sec:pointrep}). 
GoM combines  rendering quality and speed of Gaussian splatting  with geometry modeling and compatibility of deformable meshes. 
Specifically, GoM employs Gaussian splats for rendering, offering flexibility in modeling rich appearances and enabling real-time performance (\secref{sec:rendering}). GoM utilizes a skeleton-driven deformable mesh, %
enabling the creation of compact, topologically complete digital avatars, while easing mesh articulation %
through forward kinematics (\secref{sec:articulate}). Crucially, to integrate both  representations, we attach a Gaussian to each mesh face. This method differs from traditional mesh techniques that rely on texturing or vertex coloring to enhance rendering. It also differs from standard freeform Gaussian splats, thereby better regularizing Gaussians for novel poses.
Furthermore, to tackle view dependency, we factorize the final RGB color into a pseudo albedo map rendering and a pseudo shading map prediction.
This entire representation can be inferred from a single input video without additional training data (\secref{sec:training}). 
We find this dual representation to balance performance and efficiency effectively. Importantly, the entire animation and rendering of GoM are fully compatible with graphics engines, such as OpenGL.

\begin{figure}[t]
    \centering
    \includegraphics[width=\linewidth]{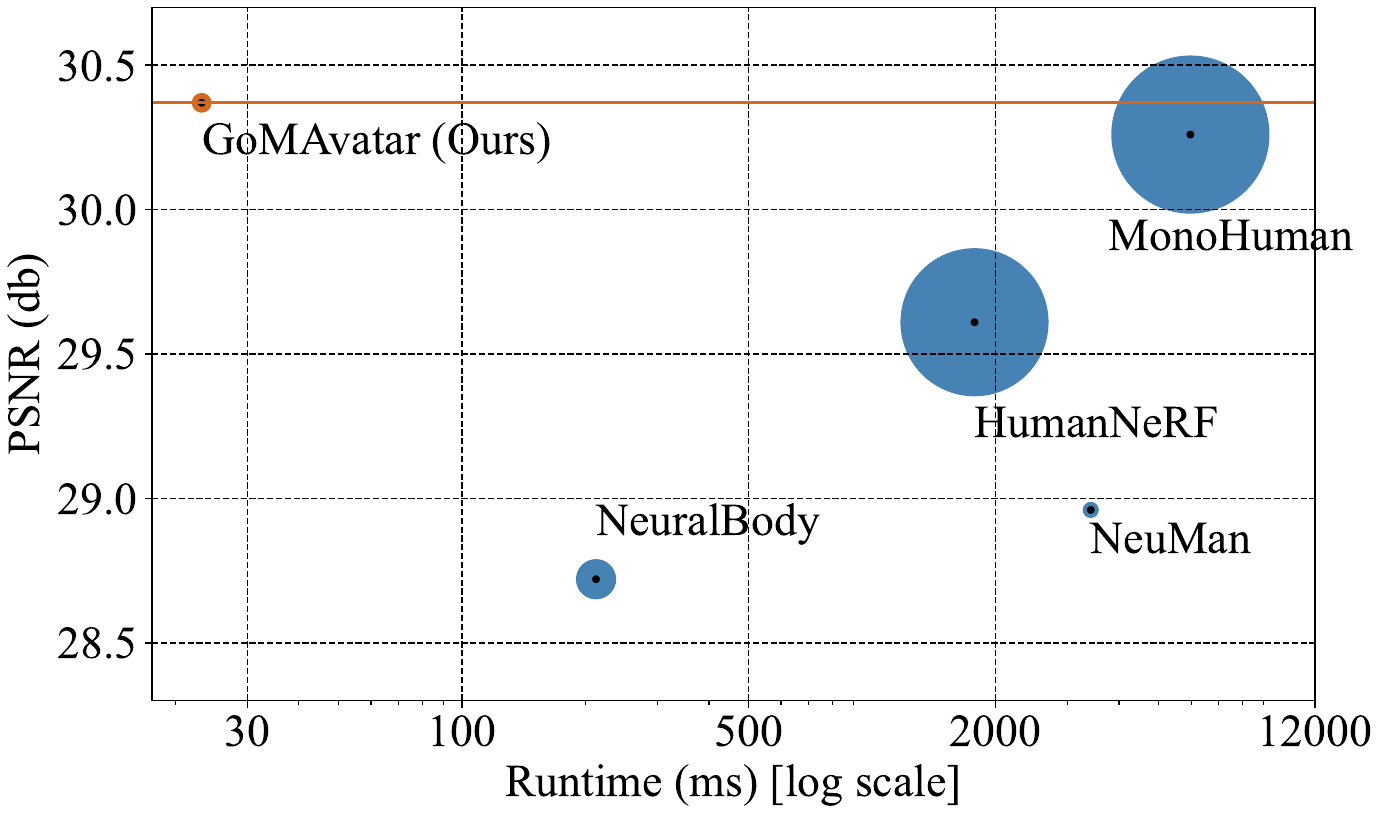}
    \vspace{-0.7cm}
    \caption{Our approach is simultaneously faster (represented by $x$ coordinates of circle centers , smaller is better), memory-efficient (represented by circle size, smaller is better), and renders at a higher quality (represented by $y$ coordinates of circle centers, higher is better). The horizontal brown line denotes our PSNR. 
    }
    \label{fig:performance}
    \vspace{-0.6cm}
\end{figure}

We conducted extensive experiments on the ZJU-MoCap data~\cite{peng2021neural}, PeopleSnapshot~\cite{alldieck2018video} and YouTube videos. \method~matches or surpasses the rendering quality of the best monocular human modeling algorithms (\method~reaches {\bf 30.37} dB PSNR in novel view synthesis and {\bf 30.31} dB PSNR in novel pose synthesis). Meanwhile, it is faster than competing algorithms, reaching a rendering speed of {\bf 43} FPS on an NVIDIA A100 GPU and remains compact in memory, only costing {\bf 3.63} MB per subject (\figref{fig:performance}). To summarize, our main contributions are:
\begin{itemize}
    \item  We introduce the Gaussians-on-Mesh representation for efficient, high-fidelity articulated human reconstruction from a single video, combining Gaussian splats with deformable meshes for real-time, free-viewpoint rendering.
    \item We design a unique differentiable shading module for view dependency, splitting color into a pseudo albedo map from Gaussian splatting and a pseudo shading map derived from the normal map.
\end{itemize}

\section{Related Work}
\label{sec:related}

\noindent\textbf{Representations for novel view synthesis.}
Several representations have been proposed for the task of novel view synthesis, such as light fields~\cite{Levoy1996LightFR, Gortler1996TheL, Attal2021LearningNL}, layered representations~\cite{Shade1998LayeredDI, Shih20203DPU, Wizadwongsa2021NeXRV, Zhou2018StereoML}, voxels~\cite{Lombardi2019NeuralV, sitzmann2019deepvoxels}, and meshes~\cite{Iizuka2016LetTB, Hasselgren2022ShapeL}.
Recently, several works also demonstrated the effectiveness of an implicit representation,~\ie,~a neural network, for a scene~\cite{chen2018implicit_decoder, mescheder2019occupancy, Park_2019_CVPR}.
Further, neural radiance fields (NeRFs)~\cite{Mildenhall2020NeRFRS} utilize a volume rendering equation~\cite{Kajiya1984RayTV} to optimize the implicit representation, yielding high-quality view synthesis. Follow-up works further improve and demonstrate compelling rendering results~\cite{Zhang2020NeRFAA, Barron2021MipNeRFAM, Barron2021MipNeRF3U, Verbin2021RefNeRFSV, Barron2023ZipNeRFAG, ren-cvpr2022-nvos,MartinBrualla2020NeRFIT}.
Meanwhile, other works use volume rendering equations to optimize more explicit representations~\cite{Yu2021PlenoxelsRF, Chen2022TensoRFTR, Mller2022InstantNG}, largely accelerating the optimization procedure.
Point-based rendering (\eg, Gaussian splatting~\cite{kerbl3Dgaussians, luiten2023dynamic, wu20234d}) has recently been adopted for fast rendering. It models the scenes as a set of 3D Gaussians, each equipped with rotation, scale, and appearance-related features, and rasterizes by projecting the 3D Gaussians to the 2D image plane. To model dynamic scenes, \cite{luiten2023dynamic, wu20234d} further extend the 3D Gaussians, adding a time dependency. To regularize the 3D Gaussians through time, \cite{luiten2023dynamic} adds physically-based priors during training, and \cite{wu20234d} uses a neural network to predict the deformation of Gaussians.
Our approach is inspired by the recent progress of point-based rendering to facilitate fast rendering. More concretely, we also use Gaussian splatting for rendering. However, different from previous approaches, we propose the Gaussians-on-Mesh representation that combines 3D Gaussians with a mesh representation. By doing so, we obtain fast rendering speed as well as regularized deformation of 3D Gaussians.

\noindent\textbf{Human modeling.}
Early works to model humans rely on templates,~\eg,~SCAPE~\cite{anguelov2005scape} and SMPL~\cite{loper2015smpl}. Later,~\cite{PIFuICCV19, saito2020pifuhd,ren-redo2021, Zhao2022OccupancyPF} utilize (pixel-aligned) image features to reconstruct human geometry and appearance from a single image. However, such human modeling is not animatable.
ARCH~\cite{xu2014search, he2021arch++} and S3~\cite{yang2021s3} incorporate reanimation capabilities but they fall short in delivering high-quality rendering.
Recently, efforts on human geometry modeling exploit implicit representations~\cite{chen2021snarf, Chen2022FastSNARFAF, Mihajlovi2022COAPCA, Tiwari2021NeuralGIFNG, Jeruzalski2019NASANA}. 
Their use of 3D scans also limits their application.
To address this limitation, human modeling from videos has received a lot of attention from the community: many prior efforts utilize implicit representations and differentiable renderers for either non-animatable~\cite{peng2021neural} or animatable~\cite{Hu2021HVTRHV, Li2022TAVATA, Weng2022HumanNeRFFR, Wang2022ARAHAV, Jiang2022SelfReconSR, Peng2021AnimatableNerf, Remelli2022DrivableVA, Xu2021HNeRFNR, Zhang2022NDFND, liu2021neural, Zheng2022StructuredLR, Su2021ANeRFAN, Geng2023LearningNV, yu2023monohuman} scene-specific human modeling while other efforts focus on scene-agnostic modeling~\cite{Kwon2021NeuralHP, Zhao2021HumanNeRFEG, Kwon2023NeuralIA, Gao2022MPSNeRFG3, ren-redo2021, Chen2023GMNeRFLG, Hu2023SHERFGH, Kim2023YouOT, zhaopgdvs2023}.
In this study, our approach focuses on \textit{scene-specific} modeling following prior works. Different from the common pure implicit representations, we utilize an explicit representation termed Gaussians-on-Mesh. The explicit canonical geometry enables us to apply well-defined forward kinematics, such as linear blend skinning, to transform from the canonical space to the observation space. In contrast, methods using implicit representations can only perform mapping in a backward manner, i.e., from the observation space to the canonical space, which is inherently ill-posed and ambiguous.

\noindent\textbf{Real-time rendering of animatable human modeling.}
The key to real-time rendering in our approach is the co-design of an explicit geometry representation and rasterization: Gaussian splatting and mesh rasterization are faster than volume rendering in general.
This principle has been explored by prior efforts to accelerate the rendering of general-purpose NeRFs. Representative approaches propose to either bake~\cite{Hedman2021BakingNR, Yu2021PlenOctreesFR, bakedsdf} or cache~\cite{Garbin2021FastNeRFHN} the trained implicit representation. Another line of work exploits mesh-based rasterization to boost the inference speed~\cite{lin2022neurmips, Chen2022MobileNeRFET, bakedsdf}. 
Inspired by the success, concurrent works explore efficient NeRF rendering for humans~\cite{Rocco2023RealtimeVR, Geng2023LearningNV}.
{Note,~\cite{Rocco2023RealtimeVR}
firstly trains a NeRF representation and then bakes it into a mesh for real-time rendering. However, the second baking stage is shown to harm the rendering quality. In contrast, the proposed Gaussians-on-Mesh representation is trained end-to-end, achieving a superior quality-speed trade-off.

\section{Gaussians-on-Mesh (GoM)}
\label{sec:method}

\begin{figure}
    \centering
    \includegraphics[width=\linewidth,trim={0cm 0 0.5cm 0},clip]{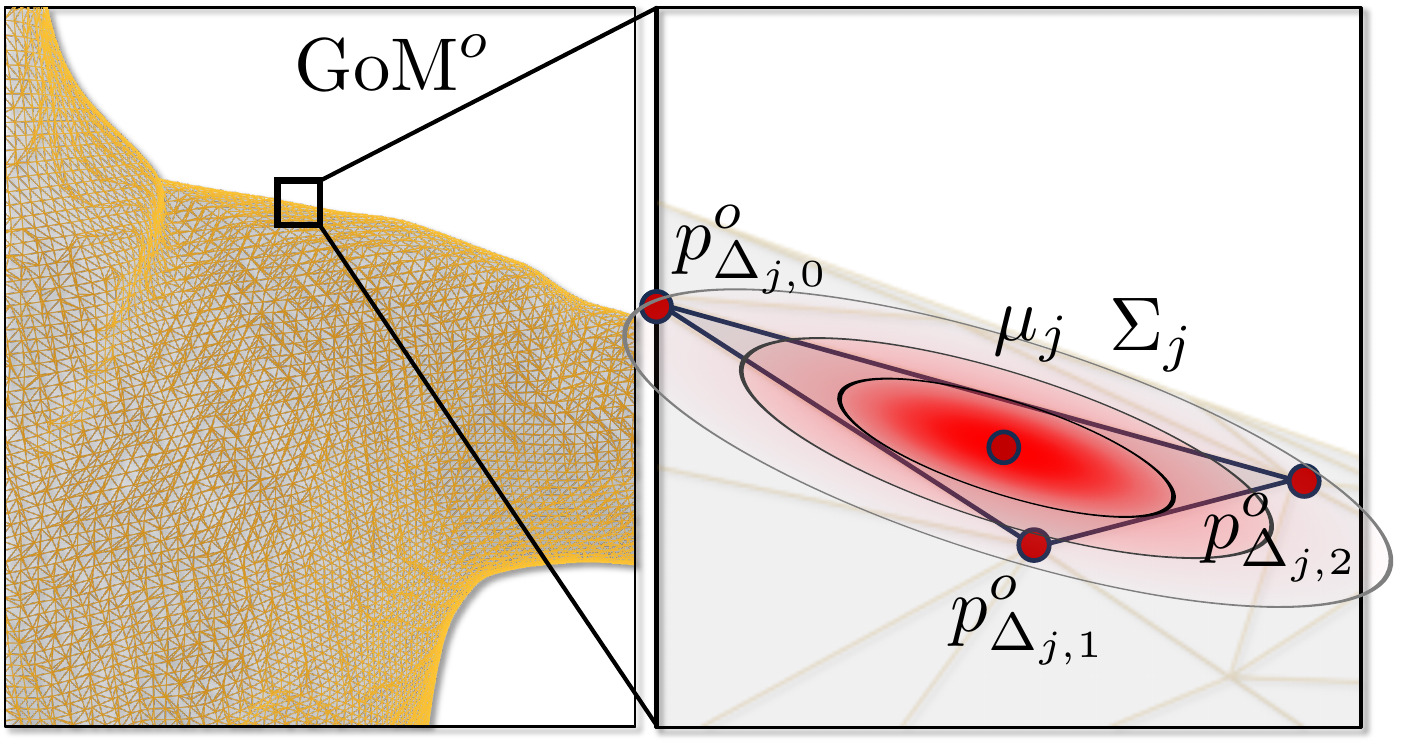}
    \vspace{-0.7cm}
    \caption{\textbf{Gaussians-on-Mesh (GoM).} We learn Gaussians in the local coordinates of each triangle and transform them to the world coordinate based on the triangle's shape. We initialize the rotation $r_{\theta,j} \in so(3)$ to zeros and scale $s_{\theta, j} \in \mathbb{R}^3$ to ones so that we start with a  Gaussian that's thin along the normal axis of the triangle. 
    Meanwhile, the projection of the ellipsoid $\{x: (x-\mu_j)^T \Sigma_j^{-1}(x-\mu_j)=1\}$ on the triangle recovers the Steiner ellipse.
    See~\secref{sec:pointrep} and the appendix for details.
    }
    \label{fig:representation}
    \vspace{-0.6cm}
\end{figure}

In the following, we present the Gaussians-on-Mesh (GoM) representation, how to render it, and its articulation. The goal of the proposed representation is to combine the benefits of both Gaussian splatting and meshes while alleviating some of their individual shortcomings. Concretely,  by using Gaussian splatting, we attain a high-quality real-time rendering capability, achieving 43 FPS. By utilizing a mesh, we conduct effective articulation 
in a forward manner while also regularizing the underlying geometry.

\noindent\textbf{Overview.} Given a monocular video capturing a human subject of interest, we aim to learn a canonical Gaussians-on-Mesh representation $\text{GoM}_\theta^c$ such that we can render that human in  \textit{real-time}  given any camera intrinsics $K \in \mathbb{R}^{3\times 3}$, extrinsics $E \in SE(3)$, and a human pose $P$. Note, here and below, parameters $\theta$ indicate that the corresponding function or variable is learnable and superscript $c$ indicates the canonical pose space. 
To render, we first articulate $\text{GoM}_\theta^c$ to the observation space to obtain
\begin{align}
    \text{GoM}^o = \texttt{Articulator}_\theta \left( \text{GoM}_\theta^c, P \right), \label{eq: articulate}
\end{align}
where $\text{GoM}^o$ denotes the Gaussians-on-Mesh representation in the observation space.
To obtain a rendering with resolution $H\times W$, we formulate a neural renderer to yield the human appearance $I \in \mathbb{R}^{H\times W \times 3}$ and the alpha mask $M \in \mathbb{R}^{H\times W \times 1}$. Formally, 
\begin{align}
    (I, M) = \texttt{Renderer}_\theta \left( K, E, \text{GoM}^o \right). \label{eq: render}
\end{align} 
The final rendering is obtained from a classical alpha-composition based on $I$ and $M$.
We will first discuss the details of the Gaussians-on-Mesh human representation in~\secref{sec:pointrep} and the rendering pipeline in~\secref{sec:rendering}. Then we introduce how to articulate the Gaussians-on-Mesh representation in~\secref{sec:articulate}.

\subsection{Gaussians-on-Mesh Representation}
\label{sec:pointrep}

The core of our approach is the Gaussians-on-Mesh (GoM) representation in the canonical space. The design of the representation is motivated by the following two key considerations:
1) GoM can  be rendered efficiently  through Gaussian splatting~\cite{kerbl3Dgaussians} which eliminates the need of dense samples along  rays used  in volume rendering; 
2) By attaching Gaussians to a mesh, we effectively adapt the shapes of Gaussians to different human poses and enable regularization.

Formally, our canonical Gaussians-on-Mesh representation is specified via a collection of points and faces with associated attributes:
\begin{align}
    \text{GoM}_\theta^c \triangleq \{\{v_{\theta,i}^c\}_{i=1}^V, \{f_{\theta,j}\}_{j=1}^F\}. \label{eq: Xc}
\end{align}
Here, $\{v_{\theta,i}^c\}_{i=1}^V$ and $\{f_{\theta,j}\}_{j=1}^F$ represent $V$ vertices and $F$ triangle faces along with their related attributes respectively.
We further define a vertex as 
\begin{align}
    v_{\theta,i}^c = (p_{\theta,i}^c, w_{i}), \label{eq: v attr}
\end{align}
where $p_{\theta,i}^c \in \mathbb{R}^3$ is the vertex coordinate and $w_i \in \mathbb{R}^J$ refers to the linear blend skinning weights with respect to $J$ joints.
We define the face as
\begin{align}
    f_{\theta,j} = (r_{\theta,j}, s_{\theta,j}, c_{\theta,j}, \{\Delta_{j,k}\}_{k=1}^3). \label{eq: face attr}
\end{align}
$r_{\theta,j} \in so(3)$ and $s_{\theta,j} \in \mathbb{R}^3$ define the rotation and scale of the \textit{local} Gaussian associated with a face. Further, $c_{\theta,j} \in \mathbb{R}^3$ is the color vector. $\{\Delta_{j,k}\}_{k=1}^3$ are the indices of the three vertices belonging to the $j$-th face, where $\Delta_{j,k} \in \{1, \dots, V\}$. Note that we associate Gaussian parameters with faces. We will delve into the derivation of the Gaussian distributions in the world coordinates for rendering in the following section.

\subsection{Rendering}
\label{sec:rendering}

In contrast to directly computing the final color as done by prior monocular human rendering works~\cite{Weng2022HumanNeRFFR, yu2023monohuman},
rendering of the Gaussians-on-Mesh representation decomposes the RGB image $I$ into the \textit{pseudo} albedo map $I_\text{GS}$ and the \textit{pseudo} shading map $S$, i.e., the final image $I$ is given by
\begin{align}
    I = I_\text{GS} \cdot S.\label{eq: decomposition}
\end{align}
Here, $I_\text{GS}$ is rendered by Gaussian splatting and $S$ is predicted from the normal map obtained from mesh rasterization. We find this combination of Gaussian splatting and mesh rasterization to better capture view-dependent shading effects than each individual approach while retaining efficiency.
We use `pseudo' because the decomposition is not perfect. Even though, we will show that the pseudo shading map encodes lighting effects to some extent.

We emphasize that rendering operates on the
GoM representation in the \emph{observation space} (see Eq.~\eqref{eq: render}), i.e., on  \begin{align}
    \text{GoM}^o\triangleq \{\{(p_{i}^o, w_{i})\}_{i=1}^V, \{(r_{\theta,j}, s_{\theta,j}, c_{\theta,j})\}_{j=1}^F\}. \label{eq: GoM obs}
\end{align}
Note, the only difference between $\text{GoM}^o$ and $\text{GoM}_\theta^c$ defined in~\equref{eq: Xc} is the use of observation space vertex coordinates $p_{i}^o$. \secref{sec:articulate} will provide more details about how to compute $p_{i}^o$ from the vertex coordinates in canonical space $p_{\theta,i}^c$.

In greater detail, Gaussian splatting is used to render the pseudo albedo map $I_\text{GS}$, specified in Eq.~\eqref{eq: decomposition}, and the subject mask $M$, specified in Eq.~\eqref{eq: render}. To obtain the pseudo shading map $S$, specified in Eq.~\eqref{eq: decomposition}, we use the normal map $N_\text{mesh}$ obtained via standard mesh rasterization. 
During training, we also use the subject mask $M_\text{mesh}$ which is obtained through the SoftRasterizer~\cite{liu2019soft}. We now discuss the computation of $I_\text{GS}$ and $S$.

\noindent\textbf{Pseudo albedo map $I_\text{GS}$ rendering.} 
We  render $I_\text{GS}$ and $M$ with Gaussian splatting given $F$ Gaussians in the world coordinate system $\{G_{j} \triangleq \mathcal{N}(\mu_{j}, \Sigma_{j})\}_{j=1}^F$ and the corresponding colors $\{c_{\theta,j}\}_{j=1}^F$ which are defined in Eq.~\eqref{eq: face attr}. $F$ indicates the number of faces.

Importantly, different from the original 3D Gaussian splatting that directly learns Gaussian parameters within the world coordinate system, we acquire these parameters within the local coordinate frame of each triangle face. 
Subsequently, we transform these local Gaussians into the world coordinate system, taking into account the deformations of the individual faces. This distinctive formulation allows our Gaussian representation to dynamically adapt to the varying shapes of triangles, which can change across different human poses due to articulation. 
Concretely, given a face and its local parameters $f_{\theta,j} = (r_{\theta,j}, s_{\theta,j}, c_{\theta,j}, \{\Delta_{j,k}\}_{k=1}^3)$, the mean $\mu_{j}$ of a Gaussian in  world coordinates  is the centroid of the face, i.e.,
\begin{align}
    \mu_{j} = \frac{1}{3}\sum_{k=1}^3 p_{ \Delta_{j,k}}^o \label{eq: mean}.
\end{align}
$p_{\Delta_{j,k}}^o$ is the coordinate of the triangle's vertex.
The Gaussian's covariance is
\begin{align}
    \Sigma_{j} = A_{j} (R_{j} S_{j} S_{j}^T R_{j}^T) A_{j}^T. \label{eq: cov}
\end{align}
$R_{j}$ and $S_{j}$ are the matrices encoding rotation $r_{\theta, j}$ and scale $s_{\theta,j}$. $A_{j}$ is the transformation matrix from local coordinates to world coordinates which is a function of the face vertices, i.e., $A_{j}=T(\{p_{\Delta_{j,k}}^o\}_{k=1}^3)$. We provide a detailed derivation of $A_j$  in the supplementary material. 
Through Eq.~\eqref{eq: mean} and~\eqref{eq: cov},  Gaussians are dynamically adapted to the shapes of triangles of different human poses.

\noindent\textbf{Pseudo shading map $S$ prediction.} For view-dependent shading effects, we predict the pseudo shading map from the mesh rasterized normal map $N_\text{mesh}$ via
    \begin{equation}    
        \mathbb{R}^{H \times W \times 1} \ni S = \texttt{Shading}_\theta \left(\gamma(N_\text{mesh}) \right). \label{eq: shading}
    \end{equation}
Here $\gamma(\cdot)$ denotes the positional encoding~\cite{Mildenhall2020NeRFRS}. $\texttt{Shading}_\theta$ is a $1 \times 1$ convolutional network that maps each pixel to a scaling factor.

\subsection{Articulation}
\label{sec:articulate}

Different from NeRF-based approaches~\cite{Weng2022HumanNeRFFR, yu2023monohuman, Geng2023LearningNV} that require the ill-posed backward mapping from observation space to canonical space, our articulation follows the mesh's \textit{forward} articulation, i.e., from canonical space to observation space, taking advantage of our Gaussians-on-Mesh representation.

The goal of the articulator defined in \equref{eq: articulate} is to obtain the Gaussians-on-Mesh representation in observation space, i.e., $\text{GoM}^o$ (see \equref{eq: GoM obs}),
given the canonical representation $\text{GoM}_\theta^c$ and a human pose $P$. Note, we only transform $p_{\theta,i}^c$ to $p_{i}^o$ as all the other attributes are shared.

To transform, linear blend skinning (LBS) is applied to warp the vertices to the observation space. For pose-dependent non-rigid motion, we utilize a non-rigid motion module to deform the canonical vertices before applying LBS. We refer to the space after non-rigid deformation as `the non-rigidly transformed canonical space'.

\noindent\textbf{Linear blend skinning.} We adhere to the standard linear blend skinning for the transformation of vertices from the non-rigidly transformed canonical space into the observation space as $\mathbb{R}^3 \ni p_{i}^\text{o} =$
\begin{align}
     \texttt{LBS} \left( p_{i}^\text{nr}, w_{i}, P \right) = \frac{\sum_{j=1}^J w_{i}^j (R_j^p p_{i}^\text{nr} + t_j^p)}{\sum_{k=1}^{J} w_{i}^k}. \label{eq: lbs}
\end{align}
In this equation, the human pose $P=\{(R_j^p, t_j^p)\}_{j=1}^J$ is represented by the rotations and translations of $J$ joints.
Each vertex is associated with LBS weights denoted as $w_i$. And $p_{i}^\text{nr}$ represents the coordinates in the non-rigidly transformed canonical space, which we will elaborate on next.

\noindent\textbf{Non-rigid deformation.} To transform to the non-rigidly transformed canonical space, we model a pose-dependent non-rigid deformation before LBS. Specifically, we predict an offset and add it to the $i$-th canonical vertex, i.e., 
\begin{align}
    p_{i}^\text{nr} = p_{\theta, i}^{c} + \texttt{NRDeformer}_\theta \left( \gamma (p_{\theta, i}^c), P \right). \label{eq: non rigid}
\end{align}
\texttt{NRDeformer} refers to an MLP network. $\gamma(\cdot)$ denotes the sinusoidal positional encoding~\cite{Mildenhall2020NeRFRS}.

\subsection{Pose Refinement}\label{sec: pose refine}
Human poses are typically estimated from the image and hence often inaccurate. Therefore, we follow HumanNeRF~\cite{Weng2022HumanNeRFFR} to add a pose refinement module that learns to correct the estimated poses. 
Specifically, given a human pose $\hat P=\{(\hat R_j^p, t_j^p)\}_{j=1}^J$ estimated from a video frame, we predict a correction to the joint rotations via 
\begin{align}
    \{\xi_j\}_{j=1}^J = \texttt{PoseRefiner}_\theta \left( \{\hat R_j^p\}_{j=1}^J \right). \label{eq: pose refine}
\end{align}
where $\xi_j \in SO(3)$. We obtain the updated pose $P = \{( R_j^p, t_j^p)\}_{j=1}^J = \{(\hat R_j^p \cdot \xi_j, \, t_j^p)\}_{j=1}^J$, which is used in~\equref{eq: lbs} and~\eqref{eq: non rigid}. 

It's important to note that pose refinement occurs only during novel view synthesis and the training stage to compensate for the inaccuracies in pose estimation from the videos. It is not needed for animation. 

\subsection{Training}
\label{sec:training}

We supervise the predicted RGB image $I$ and subject mask $M$ with ground-truth  $I_\text{gt}$ and $M_\text{gt}$. Our overall loss is
\begin{align}
    L = L_I + \alpha_\text{lpips} L_\text{lpips} + \alpha_M L_M + \alpha_\text{reg} L_\text{reg}.  \label{eq: loss}
\end{align} 
Here, $\alpha_*$ are weights for losses.
$L_I$ and $L_M$ are the L1 loss on the RGB images and subject masks respectively. $L_\text{lpips}$ is the LPIPS loss~\cite{zhang2018unreasonable} between predicted RGB image $I$ and ground-truth $I_\text{gt}$. We add additional regularization on the underlying mesh via  $L_\text{reg} =$
\begin{align}
     L_\text{mask} + \alpha_\text{lap} L_\text{lap} + \alpha_\text{normal} L_\text{normal} + \alpha_\text{color} L_\text{color}. \label{eq: loss reg}
\end{align}
$L_\text{mask} = \|M_\text{mesh} - M_\text{gt}\|$ is the regularization on the mesh silhouette. $L_\text{lap}=\frac{1}{N}\sum_{i=1}^N\|\delta_i\|^2$ is the Laplacian smoothing loss, where $\delta_i$ is the Laplacian coordinate of the $i$-th vertex. 
$L_\text{normal}$ is the normal consistency loss that maximizes the cosine similarity of adjacent face normals.
Similar to the normal consistency, we apply a color smoothness loss denoted as $L_\text{color}$, which penalizes the differences in colors between two adjacent faces. 

We initialize the vertices and faces with SMPL~\cite{loper2015smpl}.
We initialize the $r_{\theta,j}$ and $s_{\theta, j}$ in~\equref{eq: face attr} to zeros and ones respectively
so that we start with a thin Gaussian whose variance in the face normal axis is small. Meanwhile, the projection of the ellipsoid $\{x: (x-\mu_j)^T \Sigma_j^{-1}(x-\mu_j)=1\}$ on the triangle recovers the Steiner  ellipse (see \figref{fig:representation}).
To enhance the details, we upsample the canonical $\text{GoM}_\theta^c$ using GoM subdivision during training. We first subdivide the underlying mesh by introducing new vertices at the center of each edge, followed by replacing each face with four smaller faces. The properties of each face, as described in Eq.~\eqref{eq: face attr}, are duplicated across the newly generated faces.

\begin{table*}[!t]
\centering
\begin{adjustbox}{width=0.85\linewidth,center}
\begin{tabular}{l|rrr|rrr|r|r}
\toprule
            & \multicolumn{3}{c|}{Novel view synthesis} & \multicolumn{3}{c|}{Novel pose synthesis} & \multirow{2}{*}{\makecell{Inference \\time (ms) $\downarrow$}}  & \multirow{2}{*}{\makecell{Memory \\(MB) $\downarrow$}}  \\
            & PSNR $\uparrow$       & SSIM  $\uparrow$       & LPIPS* $\downarrow$     & PSNR $\uparrow$       & SSIM  $\uparrow$        & LPIPS*  $\downarrow$    &                &        \\
\midrule
Neural Body~\cite{peng2021neural} & 28.72       & 0.9611       & 52.25       & 28.54       & 0.9604       & 53.91       &         \cellcolor{silver} 212.3      &    16.76    \\
HumanNeRF~\cite{Weng2022HumanNeRFFR}   & 29.61       & 0.9625       & 38.45       & 29.74       & 0.9655       & 34.79       &          1776.7      &    245.73    \\
NeuMan~\cite{jiang2022neuman}      & 28.96      & 0.9479       & 60.74      & 28.75       & 0.9406       & 62.35       &       3412.5         &    \cellcolor{gold} 2.27    \\
MonoHuman~\cite{yu2023monohuman}   & \cellcolor{silver} 30.26       & \cellcolor{gold}0.9692       & \cellcolor{gold}30.92       & \cellcolor{silver} 30.05       & \cellcolor{silver} 0.9684       & \cellcolor{gold}31.51       &          5970.0      &    280.67    \\
Ours        & \cellcolor{gold}30.37	& \cellcolor{silver} 0.9689	& \cellcolor{silver} 32.53       & \cellcolor{gold}30.34 & 0\cellcolor{gold}.9688	& \cellcolor{silver} 32.39     &      \cellcolor{gold}23.2          &     \cellcolor{silver}3.63    \\
\bottomrule
\end{tabular}
\end{adjustbox}
\vspace{-0.3cm}
\caption{{\bf Quantitative results on ZJU-MoCap dataset.} Our results generally provide the best (or second best) quality across both novel view and novel pose rendering while being the fastest and having the second smallest parameter size. 
($\fboxsep=0pt\fbox{\color{gold}\rule{2.5mm}{2.5mm}}$ best, 
$\fboxsep=0pt\fbox{\color{silver}\rule{2.5mm}{2.5mm}}$ second best) 
}
\label{tab:main}
\vspace{-0.5cm}
\end{table*}

\begin{table}[]
\centering
\begin{adjustbox}{width=0.65\linewidth,center}
\begin{tabular}{l|r|r}
\toprule
            & CD $\downarrow$ & NC $\uparrow$ \\
\midrule
Neural Body~\cite{peng2021neural}&  5.1473 & 0.4985 \\
HumanNeRF~\cite{Weng2022HumanNeRFFR}   &  \cellcolor{silver} 2.8029  & 0.5039 \\
MonoHuman~\cite{yu2023monohuman}   &   \cellcolor{gold} 2.6303  &  \cellcolor{silver} 0.5205 \\
Ours        &  2.8364 & \cellcolor{gold} 0.6201 \\
\bottomrule
\end{tabular}
\end{adjustbox}
\vspace{-0.3cm}
\caption{{\bf Geometry quality evaluation.} Our approach provides the best normal consistency across all methods, and MonoHuman achieves best quality in surface geometry. ($\fboxsep=0pt\fbox{\color{gold}\rule{2.5mm}{2.5mm}}$ best, 
$\fboxsep=0pt\fbox{\color{silver}\rule{2.5mm}{2.5mm}}$ second best)}
\label{tab:geometry}
\vspace{-0.4cm}
\end{table}

\begin{table}[]
\centering
\begin{adjustbox}{width=\linewidth,center}
\begin{tabular}{l|rrr|r}
\toprule
& \multicolumn{3}{c|}{Novel view synthesis} & \multirow{2}{*}{\makecell{Inference \\time (ms) $\downarrow$}}  \\
                      & PSNR $\uparrow$ & SSIM $\uparrow$ & LPIPS $\downarrow$ & \\ \midrule
Anim-NeRF~\cite{Chen2021AnimatableNR}            & \cellcolor{silver}28.89                    & 0.9682                   & \cellcolor{gold}0.0206 & 217.00\\
InstantAvatar~\cite{jiang2023instantavatar}         & 28.61                    & \cellcolor{silver}0.9698                  & 0.0242 & \cellcolor{silver}71.26\\
Ours & \cellcolor{gold}30.68                   & \cellcolor{gold}0.9767                  & \cellcolor{silver}0.0213 & \cellcolor{gold}25.82 \\ \bottomrule
\end{tabular}
\end{adjustbox}
\vspace{-0.3cm}
\caption{{\bf Quantitative results on PeopleSnapshot dataset.} Our approach provides the best results regarding PSNR and SSIM while being the fastest in inference. 
($\fboxsep=0pt\fbox{\color{gold}\rule{2.5mm}{2.5mm}}$ best, $\fboxsep=0pt\fbox{\color{silver}\rule{2.5mm}{2.5mm}}$ second best) 
}
\label{tab:main_peoplesnapshot}
\vspace{-0.7cm}
\end{table}

\section{Experiments}
\label{sec:exp}

We evaluate \method\ on the ZJU-MoCap dataset~\cite{peng2021neural}, the PeopleSnapshot dataset~\cite{alldieck2018video} and on YouTube videos, comparing with state-of-the-art human avatar modeling methods from monocular videos. We showcase our method's rendering quality under novel views and poses, as well as its speed and geometry.

\subsection{Experimental setup}
\noindent\textbf{Datasets.} We validate our proposed approach on ZJU-MoCap~\cite{peng2021neural} data, PeopleSnapshot~\cite{alldieck2018video} data and Youtube videos. \textbf{ZJU-MoCap}: The ZJU-MoCap dataset provides a comprehensive multi-camera, multi-subject benchmark for human rendering evaluation. It has 9 dynamic human videos captured by 21 synchronized cameras.
In our paper, to ensure a fair comparison, we adhere to the training/test split in MonoHuman~\cite{yu2023monohuman} and follow their monocular video human rendering setting. 
We validate our approach on six subjects (377, 386, 387, 392, 393, and 394) in the dataset. For each subject, the first 4/5 frames from Camera 0 are used for training. We use the corresponding frames in the remaining cameras to evaluate novel view synthesis, and the last 1/5 frames from all views to evaluate novel pose synthesis. \textbf{PeopleSnapshot}: The PeopleSnapshot dataset provides monocular videos where humans rotate in front of the cameras. We follow the evaluation protocol in InstantAvatar~\cite{jiang2023instantavatar} to validate our approach. We report results averaged on four subjects (f3c, f4c, m3c, and m4c) and refine the test poses.
\textbf{Youtube videos}: We qualitatively validate our approach on Youtube dancing videos used in HumanNeRF~\cite{Weng2022HumanNeRFFR}. We generate the subject masks with MediaPipe~\cite{lugaresi2019mediapipe}, and the SMPL poses with PARE~\cite{kocabas2021pare}.

\begin{figure*}
    \centering
    \vspace{-0.4cm}
    \includegraphics[width=1.0\linewidth,trim=3cm 0 3cm 0]{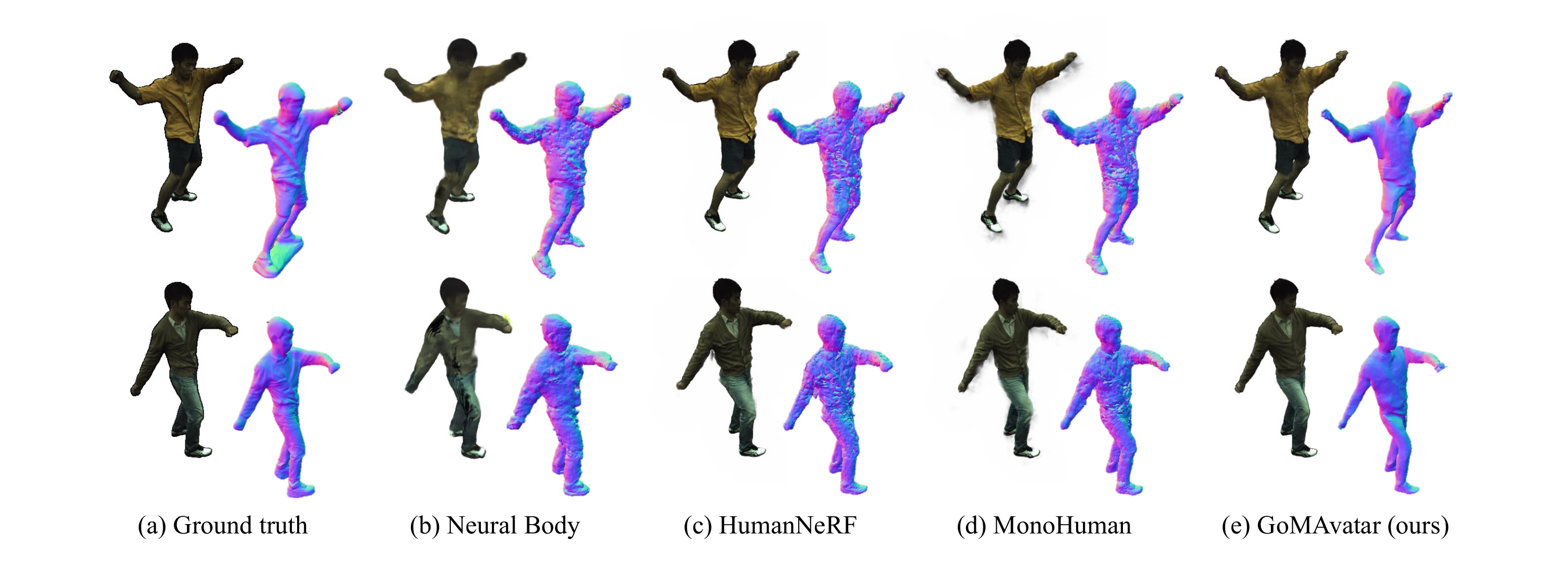}
    \vspace{-0.9cm}
    \caption{\textbf{Qualitative comparison to state-of-the-arts.} In each pair, we render the RGB image and normal map. The normal map is rendered from the extracted mesh. We show that our approach can produce realistic details in both rendered images and geometry, while other approaches struggle to generate a smooth mesh.}
    \label{fig:qualitative}
    \vspace{-0.5cm}
\end{figure*}

\noindent\textbf{Baselines.} 
We compare with state-of-the-art approaches for single-video articulated human capturing algorithms, including NeuralBody~\cite{peng2021neural}, HumanNeRF~\cite{Weng2022HumanNeRFFR}, NeuMan~\cite{jiang2022neuman}, MonoHuman~\cite{yu2023monohuman}, Anim-NeRF~\cite{Chen2021AnimatableNR} and InstantAvatar~\cite{jiang2023instantavatar}. Similar to our method, these methods take as input a single video and 3D skeleton and output an articulated neural human representation, that can facilitate both novel view and novel pose synthesis.

\noindent\textbf{Evaluation metrics.} We report PSNR, SSIM and LPIPS or LPIPS* ($=\text{LPIPS}\,\times\,1000$) for novel view synthesis and novel pose synthesis. 
To compare the geometry, we report Chamfer Distance (CD) and the Normal Consistency (NC) following the protocol in ARAH~\cite{Wang2022ARAHAV}. 
For normal consistency, we compute $1 - L2$ distance between normals for 1) each vertex in the ground-truth mesh; and 2) its closest vertex in the predicted mesh.
We also benchmark the inference speed in milliseconds (ms) / frame on an NVIDIA A100 GPU and the memory cost (the size of parameters used in inference).

\subsection{Quantitative results}

\tabref{tab:main} presents our results on ZJU-MoCap data following MonoHuman's split. In terms of perceptual performance, our approach achieves PSNR/SSIM/LPIPS* of 30.37/0.9689/32.53 on novel view synthesis and 30.34/0.9688/32.39  on novel pose synthesis, which is on par with the top-performing competitive methods MonoHuman. 
Notably, in terms of inference time, our approach achieves a rendering speed of 23.2ms/frame (43 FPS), which is 257$\times$ faster than MonoHuman, 76$\times$ faster than HumanNeRF, and more than 9$\times$ faster than any competing algorithm. These results indicate that our proposed method enables real-time articulated neural human rendering from a single video. Meanwhile, our approach is memory-efficient (3.63 MB parameters), which is smaller than all competitive methods except NeuMan~\cite{jiang2022neuman}.

We also evaluate the Chamfer distance and the normal consistency between predicted geometry and pseudo ground-truth geometry in~\tabref{tab:geometry}. Note that the pseudo ground-truths are generated from NeuS~\cite{wang2021neus} on all viewpoints and are then filtered, following ARAH~\cite{Wang2022ARAHAV}. Our approach significantly outperforms NeRF-based approaches in terms of normal consistency, which indicates that our approach can learn meaningful geometry. Note that our Chamfer distance is slightly worse than HumanNeRF and MonoHuman. It is possibly due to the use of 3D Gaussians, which have thickness in the surface normal direction. The rendered mask is larger than the actual mesh's silhouette. Hence, our meshes are a bit smaller than the `real' meshes.

Following InstantAvatar's split, we evaluate our approach on four subjects in PeopleSnapshot dataset in~\tabref{tab:main_peoplesnapshot}. Our approach achieves the PSNR/SSIM/LPIPS/inference time of 30.68/0.9767/0.0213/25.82ms, significantly outperforming InstantAvatar's 28.61/0.9698/0.0242/71.26ms. Compared to Anim-NeRF's PSNR/SSIM/LPIPS of 28.89/0.9682/0.0206, our PSNR and SSIM are significantly better, while LPIPS is on par. Also, Anim-NeRF renders at a speed of 217ms/frame on an Nvidia A100, while ours achieves 25.82ms/frame, being 8.4$\times$ faster.

\subsection{Qualitative results}
\noindent\textbf{Novel view synthesis.} 
\begin{figure}
    \centering
    \begin{adjustbox}{width=\linewidth,center}
    \setlength\tabcolsep{1pt} 
    \begin{tabular}{cccc}
    \includegraphics[width=0.22\linewidth,trim={8cm 0 6cm 0},clip]{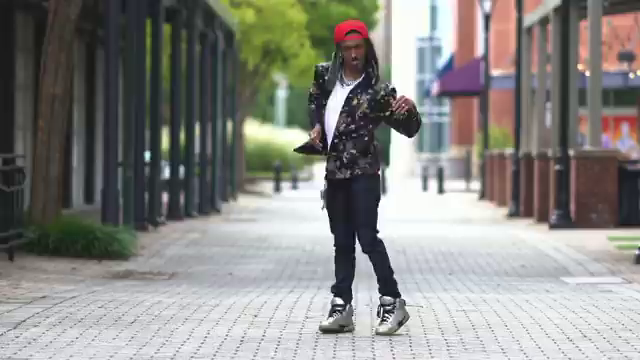}
    &
    \includegraphics[width=0.22\linewidth,trim={8cm 0 6cm 0},clip]{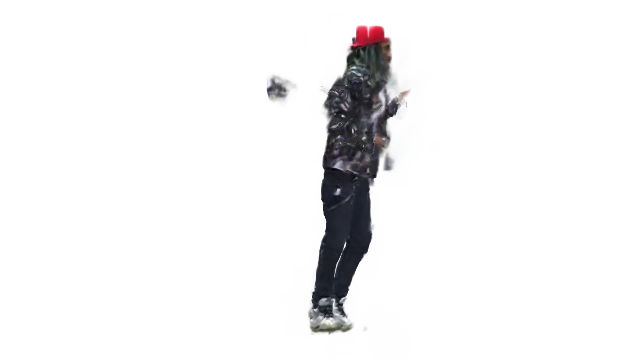}
    &
    \includegraphics[width=0.22\linewidth,trim={8cm 0 6cm 0},clip]{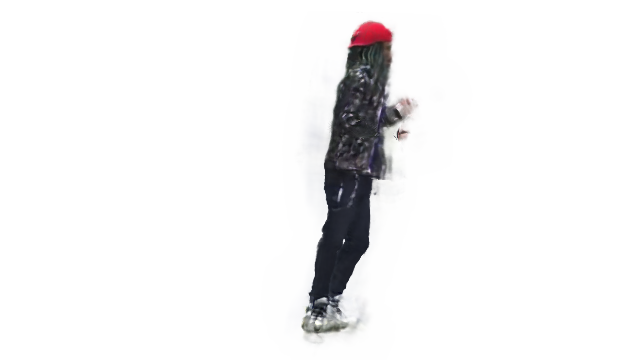}
    &
    \includegraphics[width=0.22\linewidth,trim={8cm 0 6cm 0},clip]{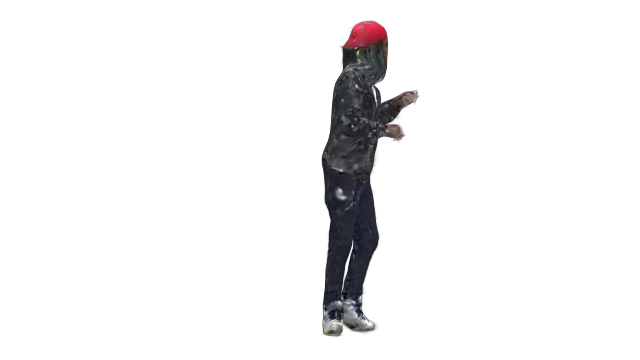}
    \\
    \includegraphics[width=0.22\linewidth,trim={4cm 0 6cm 8cm},clip]{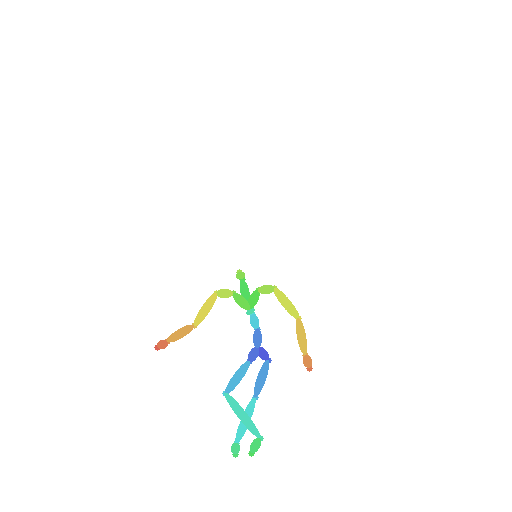}
    &
    \includegraphics[width=0.22\linewidth,trim={4cm 1.5cm 6cm 6.5cm},clip]{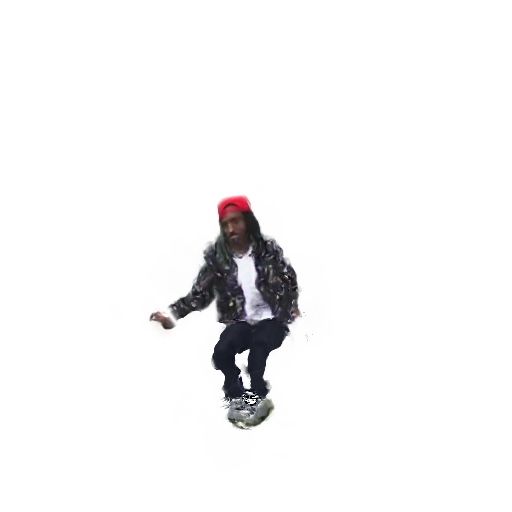}
    &
    \includegraphics[width=0.22\linewidth,trim={4cm 1.5cm 6cm 6.5cm},clip]{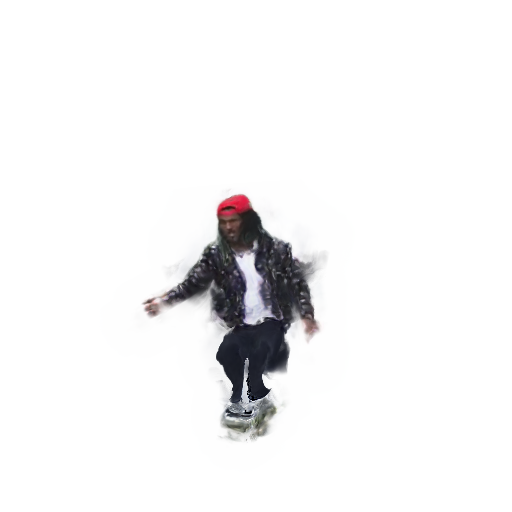}
    &
    \includegraphics[width=0.22\linewidth,trim={4cm 0 6cm 8cm},clip]{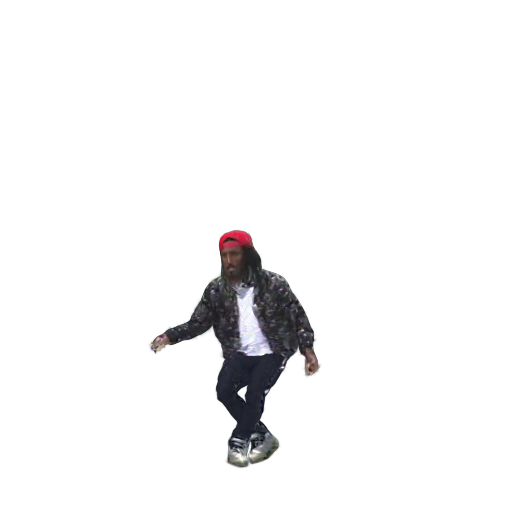}
    \\
    {\small (a) Reference} & {\small (b) HumanNeRF} & {\small (c) MonoHuman} & {\small (d) Ours}
    \end{tabular}
    \end{adjustbox}
    \vspace{-0.3cm}
    \caption{\textbf{Qualitative results on YouTube  videos.} The first image is the reference image. We compare  novel view synthesis in the first row and  novel pose synthesis in the second row.
    }
    \label{fig:in-the-wild}
    \vspace{-0.6cm}
\end{figure}
We provide a qualitative comparison with NeuralBody, HumanNeRF and MonoHuman on rendered images and normal maps in~\figref{fig:qualitative}. The normal maps are rendered from the extracted meshes. As can be seen from the figure, our approach captures fine details, such as facial features and wrinkles, and avoids the ``ghost effect'' and ``floaters'' observed in HumanNeRF's and MonoHuman's output (see the armpit of the second subject in HumanNeRF's rendering and floaters around MonoHuman's rendering). The ghost effect typically occurs when two body parts come too close, an artifact due to HumanNeRF's and MonoHuman's voxel-based inverse blend skinning. Specifically, limited by the resolution of the LBS weights, the free space is affected by two unrelated body parts and thus obtains a large foreground score.  The floaters are typical volume rendering artifacts as in other NeRF representations. In contrast, our approach uses explicit geometry and thus does not suffer from both issues.
We additionally test our approach on YouTube dancing videos in the first row of~\figref{fig:in-the-wild}. Note that the human poses and masks are predicted and thus inaccurate. However, our method  still renders novel views well, while HumanNeRF and MonoHuman suffer from imperfect masks and produce floaters.

\noindent\textbf{Novel pose synthesis.} 
\begin{figure}
    \centering
    \begin{adjustbox}{width=\linewidth,center}
    \setlength\tabcolsep{1pt} 
    \begin{tabular}{cccc}
    \includegraphics[width=0.25\linewidth,trim={6cm 1cm 4.5cm 7cm},clip]{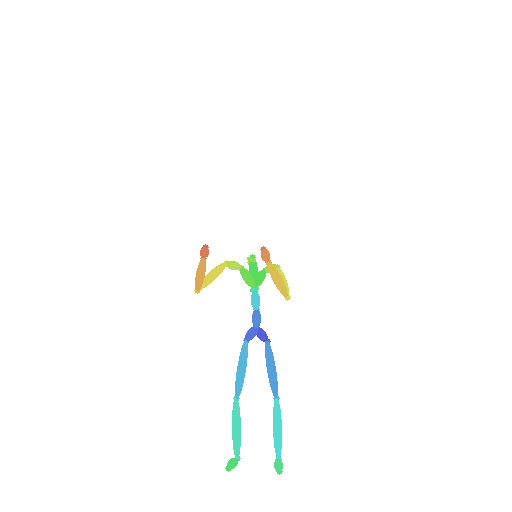}
    &
    \includegraphics[width=0.25\linewidth,trim={6cm 1cm 4.5cm 7cm},clip]{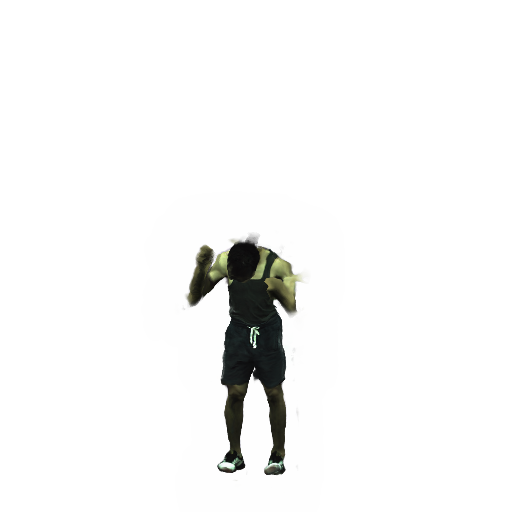}
    &
    \includegraphics[width=0.25\linewidth,trim={6cm 1cm 4.5cm 7cm},clip]{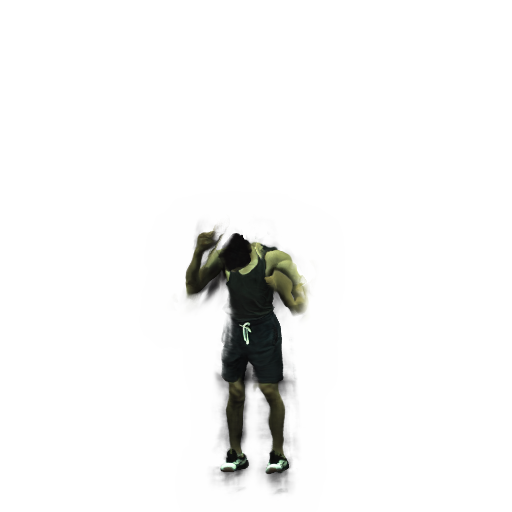}
    &
    \includegraphics[width=0.25\linewidth,trim={6cm 1cm 4.5cm 7cm},clip]{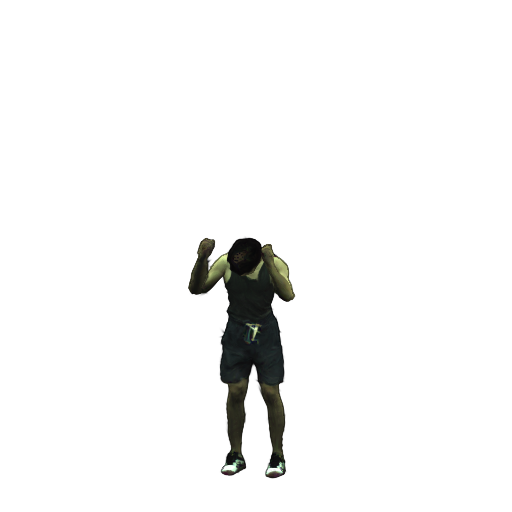}
    \\
    {\small (a) Target pose} & {\small (b) HumanNeRF} & {\small(c) MonoHuman} & {\small (d) Ours}
    \end{tabular}
    \end{adjustbox}
    \vspace{-0.3cm}
    \caption{\textbf{Novel pose synthesis}. Poses are from using poses generated from MDM~\cite{tevet2022human}.}
    \label{fig:nps}
    \vspace{-1.0cm}
\end{figure}
We render novel poses generated from MDM~\cite{tevet2022human}, as depicted in~\figref{fig:nps}. Remarkably, our approach performs effectively even in extremely challenging poses characterized by self-penetration, such as sitting. In contrast, both HumanNeRF and MonoHuman lack the capability to handle such self-penetration, due to the voxel-based inverse blend skinning (see the incomplete left hands). We validate our approach on novel view synthesis using an in-the-wild YouTube video, as illustrated in the second row of \figref{fig:in-the-wild}. Specifically, when rendering the avatar in a leg-crossing pose, both HumanNeRF and MonoHuman fail to produce accurate results, whereas our approach successfully renders the pose with fidelity.

\subsection{Ablation studies}
\label{sec: exp_ablation}
\vspace{-0.1cm}

\noindent\textbf{Canonical representation.} We conduct  ablation studies on the Gaussians-on-Mesh (GoM) representation and 3D Gaussians or meshes alone, as summarized in~\tabref{tab:representation}. In the 3D Gaussians experiment, we only use Gaussian splatting for rendering, and supervise the rendered image and subject mask during training. We initialize the Gaussians' centroids as the vertices of the canonical T-pose SMPL mesh and directly learn their centroids, rotations and scales in the world coordinates, which differs from the triangle's local coordinates  used in our approach. We also compare with just using a mesh: We initialize using the canonical SMPL mesh and attach the pseudo-albedo colors to the vertices. We render the RGB image and subject mask with mesh rasterization~\cite{liu2019soft}. We supervise the rendered image and subject mask and apply all regularizations in Eq.~\eqref{eq: loss reg}. We also utilize the color decomposition in Eq.~\eqref{eq: decomposition}.

We find 3D Gaussians alone suffer from overfitting: without geometry regularization, Gaussians are too flexible and achieve similar rendering quality on training images, while the outputs are undesirable during inference. When using only the mesh, optimization is a known challenge. In contrast, GoM alleviates these issues and combines the strengths of both representations. GoM produces the highest rendering quality among the three representations.

\noindent\textbf{Local Gaussians vs.~world Gaussians.} We compare three choices of attaching Gaussians to the mesh: 1) \textbf{World Gaussians}: We associate the Gaussian's centroid with the face's centroid (Eq.~\eqref{eq: mean}). However, we directly learn the $r_{\theta, j}$ and $s_{\theta, j}$ in the world coordinates, i.e., $\Sigma_{j}=R_{j} S_{j} S_{j}^T, R_{j}^T$, where $R_{j}$ and $S_{j}$ are the matrix encodings of $r_{\theta, j}$ and $s_{\theta, j}$; 2) \textbf{Local fixed Gaussians}: We follow Eqs.~\eqref{eq: mean} and~\eqref{eq: cov} to compute a Gaussian's mean and covariance in the world coordinates. However, $r_{\theta, j}$ and $s_{\theta, j}$ are fixed so that the variance in the normal axis is small. Meanwhile, the projection of the ellipsoid $\{x: (x-\mu_j)^T \Sigma_j^{-1}(x-\mu_j)=1\}$ on the triangle recovers the Steiner  ellipse. 3) \textbf{Local Gaussians}: We use Eqs.~\eqref{eq: mean} and~\eqref{eq: cov} to transform the Gaussians and $r_{\theta, j}$ and $s_{\theta, j}$ are free variables.

\begin{table}[]
\centering
\begin{adjustbox}{width=0.8\linewidth,center}
\begin{tabular}{cc|rrr}
\toprule
Gaussians & Mesh                   & PSNR  $\uparrow$ & SSIM $\uparrow$ & LPIPS* $\downarrow$ \\
\midrule
\checkmark & $\times$ &  30.06    &   0.9673   &    34.13    \\
$\times$ & \checkmark &  28.93   &   0.9615   &    38.11    \\
\checkmark & \checkmark              &   30.36	& 0.9690	& 33.28     \\
\bottomrule
\end{tabular}
\end{adjustbox}
\vspace{-0.3cm}
\caption{{\bf Ablations on scene representation for novel view synthesis.} Gaussians-on-Mesh  achieves the best results.}
\label{tab:representation}
\vspace{-0.6cm}
\end{table}

\begin{table}[]
\centering
\begin{adjustbox}{width=\linewidth,center}
\setlength{\tabcolsep}{0.1cm}
\begin{tabular}{l|ccc|rr}
\toprule
                   & {\small PSNR $\uparrow$} & {\small SSIM $\uparrow$} & {\small LPIPS* $\downarrow$} & {\small CD $\downarrow$} & {\small NC $\uparrow$} \\
\midrule
World Gaussians &  30.34	& 0.9689 & 33.99 &  4.3941 & 0.6223   \\
Local Fixed Gaussians &  30.27 & 0.9685 & 34.11 & 3.0898 & 0.6247   \\
Local Flex. Gaussians &   30.36	& 0.9690	& 33.28  & 3.0728 & 0.6366  \\
\midrule
w/o Shading & 30.13	& 0.9684 & 32.07 & 3.0177 & 0.6360\\
w/ Shading &  30.36	& 0.9690	& 33.28  & 3.0728 & 0.6366  \\
\midrule
w/o Subdivision &  30.36	& 0.9690	& 33.28  & 3.0728 & 0.6366  \\
w/ Subdivision & 30.37 & 0.9689 & 32.53 & 2.8364 & 0.6201 \\
\bottomrule
\end{tabular}
\end{adjustbox}
\vspace{-0.3cm}
\caption{{\bf Ablation Studies}. \textbf{Top section}: locally deformed Gaussians help improve both geometry and rendering quality. \textbf{Middle section}: our proposed shading module enhances rendering quality. \textbf{Bottom section}: subdivision significantly improves geometry.}
\label{tab:ablation}
\vspace{-0.4cm}
\end{table}

We show the comparison in the top section of~\tabref{tab:ablation}. In terms of rendering quality, world Gaussians and local Gaussians achieve similar performance. But world Gaussians tend to enlarge the scales instead of stretching the faces, so the geometry is worse. Local fixed Gaussians can produce equally good geometry, but lose rendering flexibility.

\noindent\textbf{Shading Module.} As shown in the middle section of~\tabref{tab:ablation}, without the shading module in Eq.~\eqref{eq: decomposition}—that is, by directly using \(I_\text{GS}\)  as the RGB prediction—our model achieves a PSNR of 30.13. However, with the shading module included, the PSNR increases to 30.36. We also visualize the pseudo shading map, demonstrating that our shading module  learns lighting effects, as illustrated in Fig.~\ref{fig: shading}.

\begin{figure}
    \centering
    \includegraphics[width=\linewidth]{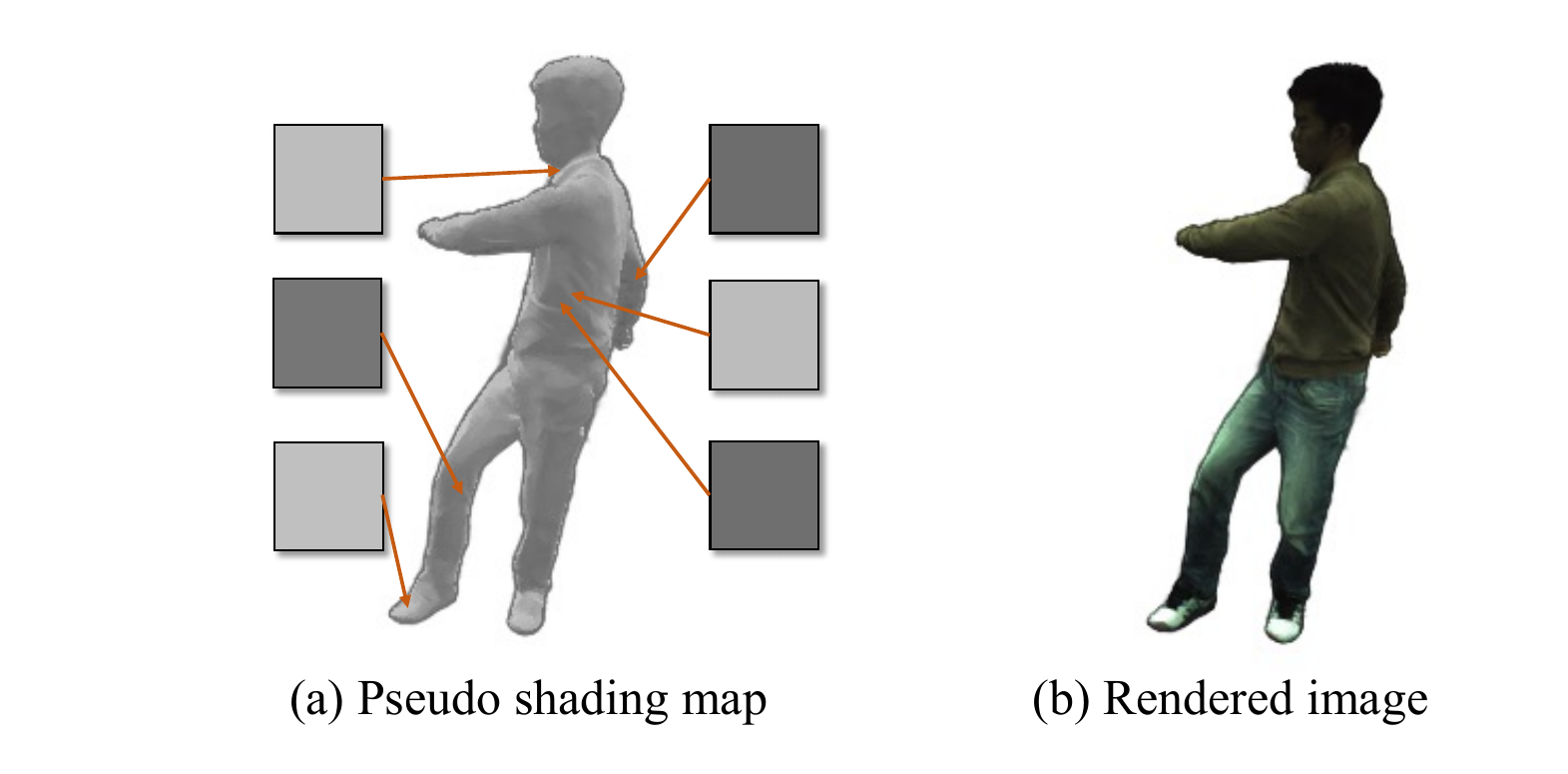}
    \vspace{-0.9cm}
    \caption{\textbf{Pseudo shading map.} We visualize the pseudo shading map and the rendered image for reference. Our approach  learns view-dependent shading effects as seen in the highlighted regions. The pseudo shading map is normalized for better visualization.}
    \label{fig: shading}
    \vspace{-0.5cm}
\end{figure}

\noindent\textbf{GoM subdivision.} We show in the bottom section of~\tabref{tab:ablation} that GoM subdivision enhances the LPIPS* from 33.28 to 32.53 and reduces the Chamfer distance from 3.0728 to 2.8364. Importantly, the geometry significantly improves with a more fine-grained mesh. Note, this increases inference time to 23.2ms per frame from 17.5ms.

\vspace{-0.3cm}
\section{Conclusion}
\vspace{-0.2cm}
\label{sec:conslusion}

We introduce \method, a framework designed for rendering high-fidelity, free-viewpoint images of a human performer, using a single input video. At the core of our method is the Gaussians-on-Mesh representation. Paired with forward articulation and neural rendering, our method renders quickly while being memory efficient. Notably, the method handles in-the-wild videos well. 

\vspace{-0.3cm}
\section*{Acknowledgement}
\vspace{-0.2cm}
Project supported by  Intel AI SRS gift,  IBM IIDAI Grant,  Insper-Illinois Innovation Grant,  NCSA Faculty Fellowship, NSF Awards \#2008387, \#2045586, \#2106825, \#2331878, \#2340254, \#2312102, and NIFA award 2020-67021-32799. We thank NCSA for providing computing resources. We thank Yiming Zuo for helpful discussions.

\clearpage
{
    \small
    \bibliographystyle{ieeenat_fullname}
    \bibliography{ref}
}

\clearpage

\setcounter{page}{1}
\maketitlesupplementary
\appendix

This supplementary material is organized as follows:
\begin{enumerate}
    \item\secref{supp sec: derivation} provides the detailed derivation of Gaussians' local-to-world transformation;
    \item\secref{supp sec: inference} details the whole inference pipeline;
    \item\secref{supp sec: architecture} shows implementation details.
    \item\secref{supp sec: additional_results} shows additional results including the quantitative results broken down per scene and additional ablation studies.
    \item\secref{supp sec: failure} showcases failure cases in our approach.
\end{enumerate}

\section{Derivation of Gaussians' local-to-world transformation}
\label{supp sec: derivation}

As described in~\secref{sec:pointrep} and~\secref{sec:rendering}, we define the rotation $r_{\theta,j}$ and the scale $s_{\theta,j}$ in the triangle's local coordinates. In order to render with Gaussian splatting, we transform the local Gaussians to the world coordinates. Specifically, the mean is transformed to the centroid of the triangle (\equref{eq: mean}) and the covariance matrix is transformed by the local-to-world transformation matrix $A_j$ (\equref{eq: cov}). We now show the detailed derivation.

Given a face and its associated local properties $f_{\theta,j} = (r_{\theta,j}, s_{\theta,j}, c_{\theta,j}, \{\Delta_{j,k}\}_{k=1}^3)$, we want to compute its Gaussian in the world coordinates $G_j = \mathcal{N}(\mu_j, \Sigma_j)$.

The Gaussian in the triangle's local coordinate is
\begin{align}
    \hat{G}_j = \mathcal{N}(\boldsymbol{0}, \hat{\Sigma}_j),~\text{where}~  \hat{\Sigma}_j=R_j S_j S_j^T R_j^T.
\end{align}
Here, $R_j$ and $S_j$ are the matrix form of $r_{\theta,j}$ and $s_{\theta,j}$ respectively. Then, we define a transformation from the triangle's local coordinates to world coordinates:
\begin{align}
    f(x) = A_j x + b_j,
\end{align}
where $A_j \in \mathbb{R}^{3 \times 3}$ and $b_j \in \mathbb{R}^3$. Therefore, the mean and covariance of the Gaussian in the world coordinates are
\begin{align}
    \mu_j & = b_j, \label{eq: supp_mu}\\
    \Sigma_j & = A_j \hat{\Sigma}_j A_j^T.
\end{align}

\paragraph{Local-to-world affine transformation $f(x)=A_j x + b_j$.} The goal of the local-to-world affine transformation is to move and reshape the Gaussian based on the location and shape of the triangle face. The world Gaussian's centroid (Eq.~\eqref{eq: supp_mu}) is put at the centroid of the triangle face, i.e.,
\begin{align}
    b_j = \frac{1}{3}\sum_{k=1}^3 p_{ \Delta_{j,k}}.
\end{align}
Here, $\{\Delta_{j,k}\}_{k=1}^3$ are the three indices of the vertices on the $j$-th triangle and hence $\{p_{ \Delta_{j,k}}\}_{k=1}^3$ are the coordinates of the vertices.

The matrix $A_j=[a_{j, 1}, a_{j, 2}, a_{j, 3}]$ takes care of the world Gaussian's shape deformation. We use $a_{j, k}$, $k\in\{1, 2, 3\}$ to denote the three columns of matrix $A_j$. Our design of $A_j$ is inspired by the Steiner ellipse of a triangle, the unique ellipse that has the maximum area of any ellipse. Specifically, we define $a_{j, 1}$ and $a_{j, 2}$ as the two semi-axes of the Steiner ellipse:
\begin{align}
    a_{j,1} &= \overrightarrow{b_j p_{ \Delta_{j,3}}} \cos t_0 + \frac{1}{\sqrt{3}} \overrightarrow{p_{ \Delta_{j,1}} p_{ \Delta_{j,2}}} \sin t_0, \\
    a_{j,2} &= \overrightarrow{b_j p_{ \Delta_{j,3}}} \cos (t_0 + \frac{\pi}{2}) + \frac{1}{\sqrt{3}} \overrightarrow{p_{ \Delta_{j,1}} p_{ \Delta_{j,2}}} \sin (t_0 + \frac{\pi}{2}),
\end{align}
where
\begin{align}
    t_0 = \frac{1}{2} \arctan \frac{\frac{2}{\sqrt{3}} \overrightarrow{b_j p_{ \Delta_{j,3}}}  \cdot \overrightarrow{p_{ \Delta_{j,1}} p_{ \Delta_{j,2}}}}{\overrightarrow{b_j p_{ \Delta_{j,3}}}^2 - \frac{1}{3}\overrightarrow{p_{ \Delta_{j,1}} p_{ \Delta_{j,2}}}^2}.
\end{align}
The third column $a_{j, 3}$ is defined along the normal vector of the triangle face:
\begin{align}
    a_{j,3} = \epsilon \cdot \text{normalize}(a_{j,1} \times a_{j, 2}).
\end{align}
We multiply the normal vector with $\epsilon$ to make sure the ellipsoid is thin along the surface normal. We set $\epsilon=1\mathrm{e}^{-3}$ in our experiments.

Given the derivation above, when the local rotation $r_{\theta,j}$ is zero and the local scale $s_{\theta,j}$ is one, i.e, $\hat{\Sigma}_j=\mathbf{I}$, the projection of the ellipsoid $\{x: (x-\mu_j)^T \Sigma_j^{-1}(x-\mu_j)=1\}$ on the triangle recovers the Steiner  ellipse, as shown in~\figref{fig:representation}.

\section{Inference Pipeline}
\label{supp sec: inference}

\begin{figure*}
    \includegraphics[width=\linewidth]{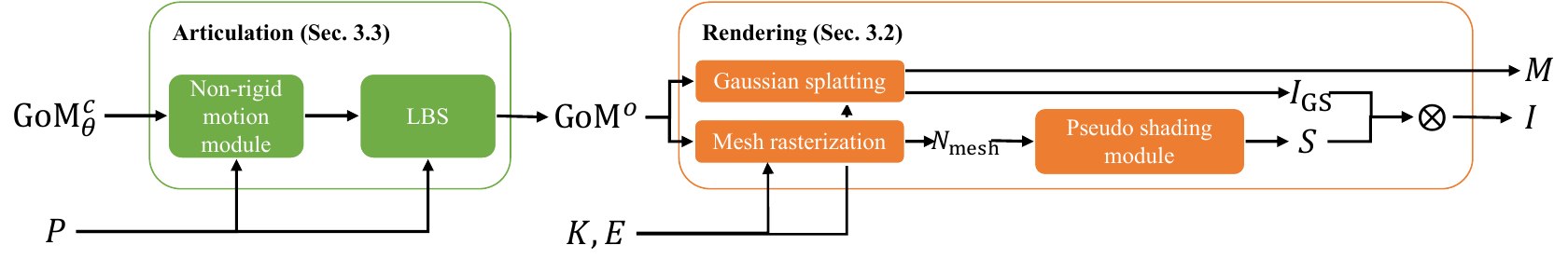}
    \vspace{-0.3cm}
    \captionof{figure}{\textbf{Inference pipeline.} Our inference pipeline has two stages: 1) \textbf{Articulation}: This stage takes the Gaussians-on-Mesh (GoM) representation in the canonical space, denoted as $\text{GoM}_\theta^c$, and the human pose $P$ as input. Utilizing the non-rigid motion module and linear blend skinning (LBS), it produces the transformed GoM representation in the observation space, referred to as $\text{GoM}^o$. 2) \textbf{Rendering}:  In this stage, the transformed $\text{GoM}^o$, along with the camera intrinsic parameters $K$ and extrinsic parameters $E$, are employed as inputs. It adopts the Gaussian splatting to generate the pseudo albedo map $I_\text{GS}$ and the subject mask $M$. Meanwhile, through mesh rasterization, it produces the normal map $N_\text{mesh}$ which is then fed into the pseudo shading module to output the pseudo shading map $S$. The final RGB image $I$ is then obtained by multiplying $I_\text{GS}$ with $S$.}
    \label{fig: supp_pipeline}
\end{figure*}

We present our inference pipeline including the modules and key inputs and outputs in~\figref{fig: supp_pipeline}.

\section{Implementation Details}
\label{supp sec: architecture}
\noindent\textbf{Architecture details.}
\textbf{1)} \texttt{GoM}$_\theta^c$ in~\equref{eq: Xc} is initialized with the SMPL mesh~\cite{loper2015smpl} under the canonical T-pose; 
\textbf{2)} $\texttt{Shading}_\theta$ in~\equref{eq: shading} is a 4-layer MLP network with 128 channels; 
\textbf{3)} $\texttt{NRDeformer}_\theta$ in~\equref{eq: non rigid} is a 7-layer MLP network with 128 channels; 
\textbf{4)} $\texttt{PoseRefiner}_\theta$ in~\equref{eq: pose refine} is a 5-layer MLP network with 256 channels. The detailed architecture of $\texttt{Shading}_\theta$, $\texttt{NRDeformer}_\theta$ and $\texttt{PoseRefiner}_\theta$ are shown in~\figref{fig: supp_arch}.

\begin{figure*}[]
    \centering
    \includegraphics[width=0.8\textwidth]{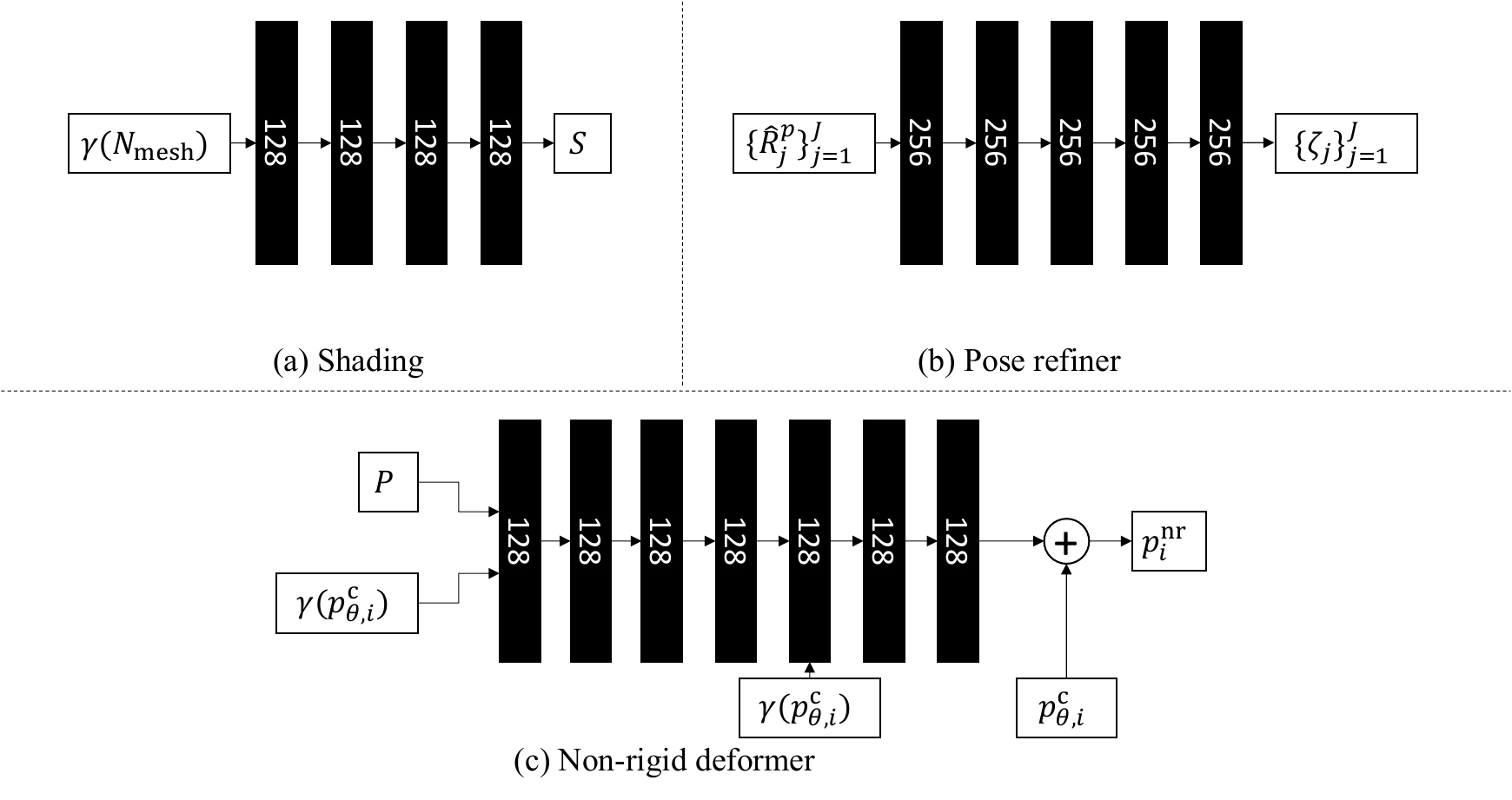}
    \vspace{-0.3cm}
    \caption{\textbf{Detailed architectures of (a) $\texttt{Shading}_\theta$, (b) $\texttt{PoseRefiner}_\theta$ and  (c) $\texttt{NRDeformer}_\theta$.}}
    \label{fig: supp_arch}
\end{figure*}

\noindent\textbf{Training details.} We use Adam optimizer~\cite{Kingma2014AdamAM} with $\beta_1=0.9$ and $\beta_2=0.999$. 
On ZJU-MoCap, We train the model for 300K iterations.
We set the learning rate of $\texttt{PoseRefiner}_\theta$ to $5\mathrm{e}{-5}$. The learning rate of the rest of the model is $5\mathrm{e}{-4}$. 
We kick off the training of $\texttt{PoseRefiner}_\theta$ and $\texttt{NRDeformer}_\theta$ after 100K and 150K iterations respectively. 
For $\texttt{NRDeformer}_\theta$, we follow HumanNeRF~\cite{Weng2022HumanNeRFFR} to adopt a HanW window during training. 
We set $\alpha_\text{lpips}=1.0$ $\alpha_M=5.0$, $\alpha_\text{reg}=1.0$ in~\equref{eq: loss}, and $\alpha_\text{lap}=10.0$, $\alpha_\text{normal}=0.1$, $\alpha_\text{color}=0.05$ in~\equref{eq: loss reg}. We subdivide the GoM after 50K iterations. On PeopleSnapshot, we train the model for 200K iterations and kick off the training of $\texttt{NRDeformer}_\theta$ after 100K iterations. We subdivide GoM once after 10K iterations. We do not refine training poses with $\texttt{PoseRefiner}_\theta$ following InstantAvatar~\cite{jiang2023instantavatar}. On in-the-wild Youtube videos, since the poses are predicted and less accurate, we kick off the training of $\texttt{PoseRefiner}_\theta$ at the start of the training process while keeping all other hyperparameters the same as ZJU-MoCap.

\section{Additional Results}
\label{supp sec: additional_results}

\subsection{Quantitative Results of Per-scene Breakdown}
\label{supp sec: breakdown}

\begin{table*}[]
\centering
\begin{tabular}{l|rrr|rrr|rrr}
\toprule
                      & \multicolumn{1}{l}{PSNR $\uparrow$} & \multicolumn{1}{l}{SSIM $\uparrow$} & \multicolumn{1}{l|}{LPIPS* $\downarrow$} & \multicolumn{1}{l}{PSNR $\uparrow$}  & \multicolumn{1}{l}{SSIM $\uparrow$} & \multicolumn{1}{l|}{LPIPS* $\downarrow$} & \multicolumn{1}{l}{PSNR $\uparrow$} & \multicolumn{1}{l}{SSIM $\uparrow$} & \multicolumn{1}{l}{LPIPS* $\downarrow$} \\ 
\midrule
\rowcolor[HTML]{EFEFEF} 
                      & \multicolumn{3}{c|}{\cellcolor[HTML]{EFEFEF}Subject 377}                          & \multicolumn{3}{c|}{\cellcolor[HTML]{EFEFEF}Subject 386}                           & \multicolumn{3}{c}{\cellcolor[HTML]{EFEFEF}Subject 387}                          \\
Neural Body           & 29.08                    & 0.9679                   & 41.17  & 29.76 & 0.9647                   & 46.96                      & 26.84                    & 0.9535                   & 60.82                      \\ 
HumanNeRF             & 29.79                    & 0.9714                   & 28.49                       & 32.10                      & 0.9642                   & 41.84                      & 28.11                    & 0.9625                   & 37.46                      \\
MonoHuman             & 30.46                    & 0.9781                   & 20.91                       & 32.99                      & 0.9756                   & 30.97                      & 28.40                    & 0.9639                   & 35.06                      \\
\method~(Ours) & 30.60                    & 0.9768                   & 23.91                       & 32.97                      & 0.9752                   & 30.36                      & 28.34                    & 0.9635                   & 36.30                      \\
\midrule
\rowcolor[HTML]{EFEFEF} 
                      & \multicolumn{3}{c|}{\cellcolor[HTML]{EFEFEF}Subject 392}                           & \multicolumn{3}{c|}{\cellcolor[HTML]{EFEFEF}Subject 393}                            & \multicolumn{3}{c}{\cellcolor[HTML]{EFEFEF}Subject 394}                          \\
Neural Body           & 29.49                    & 0.9640                   & 51.06                       & 28.50                      & 0.9591                   & 57.07                      & 28.65                    & 0.9572                   & 55.78                      \\
HumanNeRF             & 30.20                    & 0.9633                   & 40.06                       & 28.16                      & 0.9577                   & 40.85                      & 29.28                    & 0.9557                   & 41.97                      \\
MonoHuman             & 30.98                    & 0.9711                   & 30.80                       & 28.54                      & 0.9620                   & 34.97                      & 30.21                    & 0.9642                   & 32.80                      \\
\method~(Ours) & 31.04                    & 0.9708                   & 33.25                       & 28.80                      & 0.9622                   & 37.77                      & 30.44                    & 0.9646                   & 33.56        \\
\bottomrule
\end{tabular}
\vspace{-0.3cm}
\caption{\textbf{Per-scene breakdown in novel view synthesis on ZJU-MoCap dataset.}}
\label{tab: per-scene breakdown nvs}
\end{table*}

\begin{table*}[]
\centering
\begin{tabular}{l|rrr|rrr|rrr}
\toprule
                      & \multicolumn{1}{l}{PSNR $\uparrow$} & \multicolumn{1}{l}{SSIM $\uparrow$} & \multicolumn{1}{l|}{LPIPS* $\downarrow$} & \multicolumn{1}{l}{PSNR $\uparrow$}  & \multicolumn{1}{l}{SSIM $\uparrow$} & \multicolumn{1}{l|}{LPIPS* $\downarrow$} & \multicolumn{1}{l}{PSNR $\uparrow$} & \multicolumn{1}{l}{SSIM $\uparrow$} & \multicolumn{1}{l}{LPIPS* $\downarrow$} \\ \midrule
\rowcolor[HTML]{EFEFEF} 
                      & \multicolumn{3}{c|}{\cellcolor[HTML]{EFEFEF}Subject 377}                          & \multicolumn{3}{c|}{\cellcolor[HTML]{EFEFEF}Subject 386}                           & \multicolumn{3}{c}{\cellcolor[HTML]{EFEFEF}Subject 387}                          \\ 
Neural Body           & 29.29                    & 0.9693                   & 39.40  & 30.71 & 0.9661                   & 45.89                      & 26.36                    & 0.9520                   & 62.21                      \\ 
HumanNeRF             & 29.91                    & 0.9755                   & 23.87                       & 32.62                      & 0.9672                   & 39.36                      & 28.01                    & 0.9634                   & 35.27                      \\
MonoHuman             & 30.77                    & 0.9787                   & 21.67                       & 32.97                      & 0.9733                   & 32.73                      & 27.93                    & 0.9633                   & 33.45                      \\
\method~(Ours) & 30.68                    & 0.9776                   & 23.41                       & 32.86                      & 0.9737                   & 32.25                      & 28.18                    & 0.9626                   & 36.43                      \\ \midrule
\rowcolor[HTML]{EFEFEF} 
                      & \multicolumn{3}{c|}{\cellcolor[HTML]{EFEFEF}Subject 392}                           & \multicolumn{3}{c|}{\cellcolor[HTML]{EFEFEF}Subject 393}                            & \multicolumn{3}{c}{\cellcolor[HTML]{EFEFEF}Subject 394}                          \\
Neural Body           & 28.97                    & 0.9615                   & 57.03                       & 27.82                      & 0.9577                   & 59.24                      & 28.09                    & 0.9557                   & 59.66                      \\
HumanNeRF             & 30.95                    & 0.9687                   & 34.23                       & 28.43                      & 0.9609                   & 36.26                      & 28.52                    & 0.9573                   & 39.75                      \\
MonoHuman             & 31.24                    & 0.9715                   & 31.04                       & 28.46                      & 0.9622                   & 34.24                      & 28.94                    & 0.9612                   & 35.90                      \\
\method~(Ours) & 31.44                    & 0.9716                  & 33.20                       & 29.09                      & 0.9635                   & 36.02                      & 29.79                    & 0.9638                   & 33.00                  \\ \bottomrule   
\end{tabular}
\vspace{-0.3cm}
\caption{\textbf{Per-scene breakdown in novel pose synthesis on ZJU-MoCap dataset.}}
\label{tab: per-scene breakdown nps}
\end{table*}

\begin{table*}[]
\centering
\begin{tabular}{l|rrr|rrr}
\toprule
                      & \multicolumn{1}{l}{PSNR $\uparrow$} & \multicolumn{1}{l}{SSIM $\uparrow$} & LPIPS $\downarrow$  & \multicolumn{1}{l}{PSNR $\uparrow$} & \multicolumn{1}{l}{SSIM $\uparrow$} & \multicolumn{1}{l}{LPIPS $\downarrow$} \\ \midrule
\rowcolor[HTML]{EFEFEF} 
                      & \multicolumn{3}{c|}{\cellcolor[HTML]{EFEFEF}m3c}              & \multicolumn{3}{c}{\cellcolor[HTML]{EFEFEF}m4c}                                 \\
Anim-NeRF             & 29.37                    & 0.9703                   & 0.0168 & 28.37                    & 0.9605                   & 0.0268                    \\
InstantAvatar         & 29.65                    & 0.9730                   & 0.0192 & 27.97                    & 0.9649                   & 0.0346                    \\
\method~(Ours) & 31.74	& 0.9793	&0.0187&	29.78&	0.9738&	0.0282              \\  \midrule
\rowcolor[HTML]{EFEFEF} 
                      & \multicolumn{3}{c|}{\cellcolor[HTML]{EFEFEF}f3c}              & \multicolumn{3}{c}{\cellcolor[HTML]{EFEFEF}f4c}                                 \\
Anim-NeRF             & 28.91                    & 0.9743                   & 0.0215 & 28.90                    & 0.9678                   & 0.0174                    \\
InstantAvatar         & 27.90                    & 0.9722                   & 0.0249 & 28.92                    & 0.9692                   & 0.0180                    \\
\method~(Ours) & 29.83	& 0.9758 & 0.0209 & 31.38 & 0.9780 & 0.0174         \\ \bottomrule
\end{tabular}
\caption{\textbf{Per-scene breakdown in novel view synthesis on PeopleSnapshot dataset.}}
\label{tab: per-scene breakdown ps}
\end{table*}

We show the per-scene PSNR, SSIM and LPIPS* on the ZJU-MoCap dataset in Tab.~\ref{tab: per-scene breakdown nvs} and Tab.~\ref{tab: per-scene breakdown nps}. The per-scene breakdown results on PeopleSnapshot is shown in Tab.~\ref{tab: per-scene breakdown ps}.

\subsection{Qualitative Results on PeopleSnapshot}

\begin{figure}
    \centering
    \includegraphics[width=\linewidth]{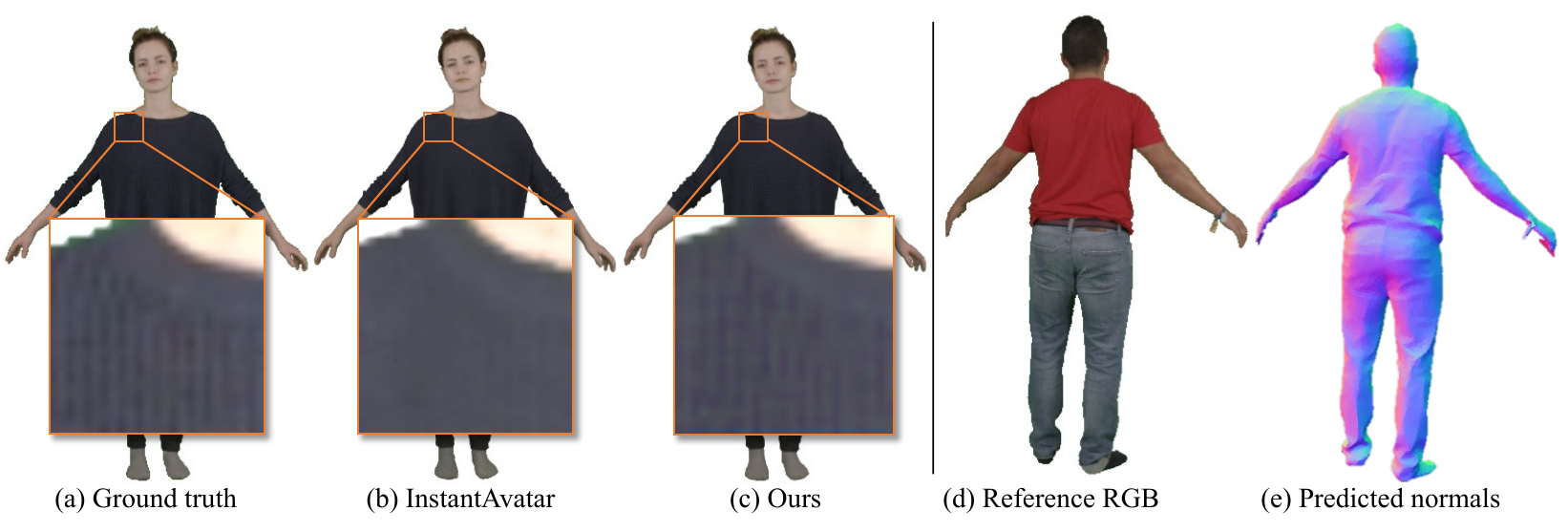}
    \caption{{\bf Qualitative results on PeopleSnapshot dataset.} On the left side, we conduct a qualitative comparison to InstantAvatar. We also show the geometry by rendering the surface normals on the right side.}
    \label{fig: rebuttal_snapshot}
\end{figure}

We conduct a qualitative comparison on PeopleSnapshot dataset in Fig.~\ref{fig: rebuttal_snapshot}. As shown below, we better capture textures better compared to InstantAvatar. Meanwhile, our approach can capture fine details in geometry, such as wrinkles.

\subsection{Sensitivity to SMPL Accuracy}

Our approach takes the human poses in the input frames as inputs. The human poses are provided in ZJU-MoCap dataset and PeopleSnapshot dataset, while we predict the poses with PARE~\cite{kocabas2021pare} for in-the-wild Youtube videos.

\begin{figure}
    \centering
    \includegraphics[width=\linewidth]{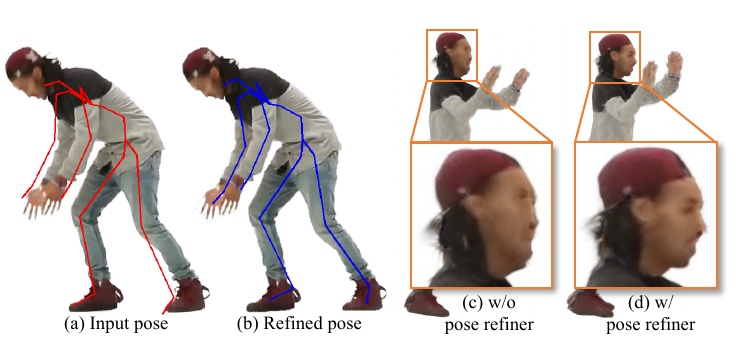}
    \caption{{\bf Robustness to SMPL accuracy.} Our method is robust to SMPL prediction. 
This can be seen in in-the-wild videos. (a)  Predicted poses have errors. (b) Pose refinement improves erroneous SMPL poses, which is crucial for in-the-wild videos (c, d).}
    \label{fig: rebuttal_smpl}
\end{figure}

The robustness to SMPL prediction can be seen in in-the-wild videos. In Fig.~\ref{fig: rebuttal_smpl}(a), we show that there are errors in pose prediction in in-the-wild videos. However, the pose refinement improves erroneous SMPL poses (Fig.~\ref{fig: rebuttal_smpl}(b)). The pose refinement is crucial for rendering in in-the-wild videos, which can be seen in Fig.~\ref{fig: rebuttal_smpl}(c, d). Without the pose refinement, the approach fails to render the correct shape of the human face. This issue is solved when equipped with the pose refinement.

\begin{table}[]
    \centering
    \begin{adjustbox}{width=\linewidth}
    \begin{tabular}{c|rrr|rr}
    \toprule
    & PSNR $\uparrow$& SSIM $\uparrow$& LPIPS*$\downarrow$ & CD $\downarrow$& NC $\uparrow$\\
    \midrule
      Original   & 30.37 & 0.9689 & 32.53 & 2.8364 &0.6201 \\
      Refined   & 30.86 & 0.9709 & 30.91 & 2.3377 & 0.6307 \\ \bottomrule
    \end{tabular}
    \end{adjustbox}
    \caption{{\bf Quantitative evaluation about sensitivity to SMPL accuracy.} We test our approach with two versions of SMPL poses on ZJU-MoCap dataset. ``Original'' refers to the poses provided in the original ZJU-MoCap dataset, which is less accurate. ``Refined'' refers to the improved version from InstantNVR~\cite{Geng2023LearningNV}.}
    \label{tab: supp_smpl}
\end{table}

In Tab.~\ref{tab: supp_smpl}, we quantitatively assess sensitivity to SMPL accuracy on ZJU-MoCap, comparing the original less-accurate SMPL poses to refined versions from InstantNVR~\cite{Geng2023LearningNV}. Hence, refining  SMPL pose improves rendering and geometry quality, but our method will {\it not} fail without.

\subsection{Ablation on Canonical Representations}

\begin{figure}
    \centering
    \includegraphics[width=0.5\linewidth]{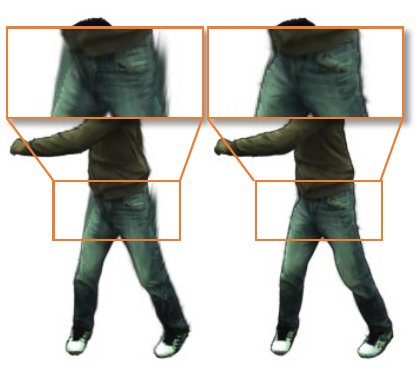}
    \caption{{\bf Qualitative comparison between GoM and Gaussians only.}}
    \label{fig: rebuttal_gaussian}
\end{figure}

In Sec.\ref{sec: exp_ablation}, we conduct a quantitative comparison of the GoM presentation and 3D Gaussians alone. Here, we show the comparison to {\it Gaussians only} qualitatively in Fig.~\ref{fig: rebuttal_gaussian}. {\it Gaussians only} yield severe artifacts on the boundary while our method attains a sharp boundary.

\section{Failure Cases}
\label{supp sec: failure}

\begin{figure}[t]
    \centering
    \begin{adjustbox}{width=\linewidth,center}
    \begin{tabular}{cc}
    \includegraphics[width=0.5\linewidth,trim={1cm 2cm 3cm 2cm},clip]{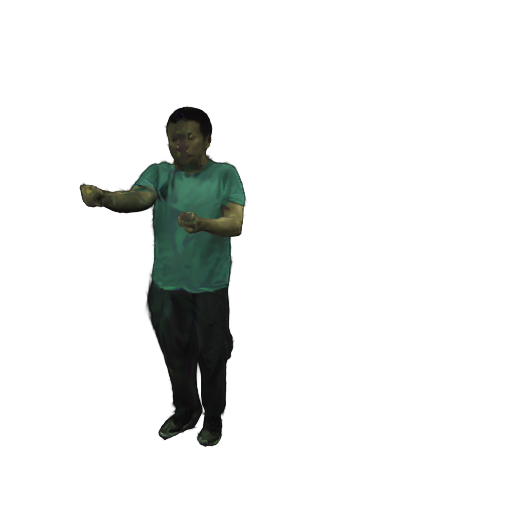}
    & 
    \includegraphics[width=0.5\linewidth,trim={1cm 3cm 3cm 1cm},clip]{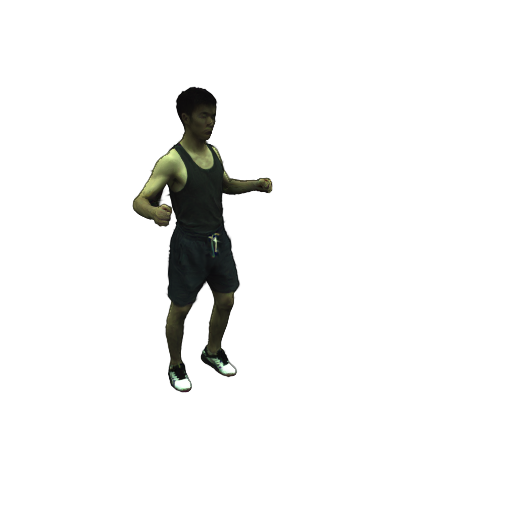}
    \\
    {\small (a)} & {\small (b)}
    \end{tabular}
    \end{adjustbox}
    \caption{\textbf{Failure cases.}}
    \label{fig: supp_failure}
\end{figure}

\begin{figure}[t]
    \centering
    \includegraphics[width=0.4\linewidth]{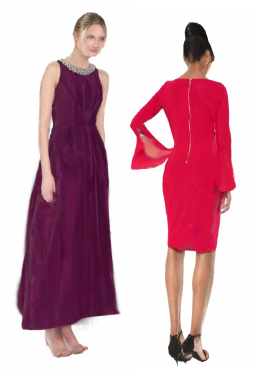}
    \caption{\textbf{Novel view synthesis on subjects in dresses.}}
    \label{fig: rebuttal_loose}
\end{figure}

We present two failure cases of our approach:
\begin{enumerate}
    \item Our approach, along with other state-of-the-art methods optimized on a per-scene basis, lacks the ability to hallucinate unseen regions. This limitation can be observed in the failure for Subject 386 in the ZJU-MoCap dataset, as shown in~\figref{fig: supp_failure}(a). In subject 386, the training frames do not cover the front view of the person. Consequently, all methods fail to generate a valid rendering from this unobserved perspective.
    \item As we associate Gaussians with the mesh in the Gaussians-on-Mesh representation, we sometimes cannot handle significant topology changes. One example is the white belt on the shorts in Subject 377 (\figref{fig: supp_failure}(b)), which dynamically shifts with the person's movement. Interestingly, when fitting clothes with different topologies from SMPL, such as dresses, our model can self-deform to fit the shapes and yield plausible novel-view renderings, even though the topology does not change. This can be seen in Fig.~\ref{fig: rebuttal_loose}. Addressing topology changes may require a pose-dependent topology update, which we leave to future work.
\end{enumerate}

\end{document}